\documentclass{article}

\PassOptionsToPackage{numbers, compress}{natbib}
\usepackage[final]{neurips_2024}

\usepackage[utf8]{inputenc} 
\usepackage[T1]{fontenc}    
\usepackage[hidelinks]{hyperref}       
\usepackage{url}            
\usepackage{booktabs}       
\usepackage{amsfonts}       
\usepackage{nicefrac}       
\usepackage{microtype}      
\usepackage{xcolor}         

\usepackage{amsmath}
\usepackage{amssymb}
\usepackage{mathtools}
\usepackage{amsthm}
\usepackage{xspace}
\newcommand{\ddlcxr}{\texttt{DDL-CXR}\xspace}
\usepackage[capitalize]{cleveref}
\usepackage{makecell}
\usepackage{marvosym}
\usepackage{wrapfig}
\usepackage{caption}
\usepackage{array}
\usepackage{enumitem}
\usepackage{arydshln}  
\usepackage[export]{adjustbox}

\title{Addressing Asynchronicity in Clinical Multimodal Fusion via Individualized Chest X-ray Generation}

\author{%
    Wenfang Yao\textsuperscript{\rm 1}\thanks{These authors contributed equally.},\ 
    Chen Liu\textsuperscript{ \rm 1,\rm 3}\footnotemark[1], \
    Kejing Yin \textsuperscript{\rm 2}\thanks{Correspondence to: Kejing Yin <cskjyin@comp.hkbu.edu.hk>}, \ 
    William K. Cheung\textsuperscript{\rm 2},\ 
    Jing Qin\textsuperscript{\rm 1}\\
  \textsuperscript{\rm 1}School of Nursing, The Hong Kong Polytechnic University\\
  \textsuperscript{\rm 2}Department of Computer Science, Hong Kong Baptist University\\
  \textsuperscript{\rm 3}School of Software Engineering, South China University of Technology
}
\begin{document}

\maketitle
\begin{abstract}
  Integrating multi-modal clinical data, such as electronic health records (EHR) and chest X-ray images (CXR), is particularly beneficial for clinical prediction tasks. However, in a temporal setting, multi-modal data are often inherently asynchronous. EHR can be continuously collected but CXR is generally taken with a much longer interval due to its high cost and radiation dose. When clinical prediction is needed, the last available CXR image might have been outdated, leading to suboptimal predictions. To address this challenge, we propose DDL-CXR, a method that dynamically generates an up-to-date latent representation of the individualized CXR images. Our approach leverages latent diffusion models for patient-specific generation strategically conditioned on a previous CXR image and EHR time series, providing information regarding anatomical structures and disease progressions, respectively. In this way, the interaction across modalities could be better captured by the latent CXR generation process, ultimately improving the prediction performance. Experiments using MIMIC datasets show that the proposed model could effectively address asynchronicity in multimodal fusion and consistently outperform existing methods.
\end{abstract}

\section{Introduction}
Clinical data in modern healthcare is documented through various complementary modalities~\cite{aljondi2020diagnostic,EHRimagefusingreview}. Electronic health records (EHRs), for instance, systematically record the progression of diseases over time, including medical histories, laboratory test results, and treatment outcomes~\cite{yin2024patnet,yin2022learning,yin2020logpar,song2019medical}. In parallel, medical imaging, such as chest X-rays (CXRs), is valuable for providing visual insights into the patient's internal anatomy, organ functions, and potential abnormalities~\cite{huang2020fusion}. Recent studies have shown that strategic integration of multimodal clinical data could lead to improved performance for clinical predictions compared to relying solely on uni-modal data~\cite{huang2020multimodal,daft-polsterl2021combining,mlhc2022hayatmedfuse,stahlschmidt2022multimodal,edward2023,yao2024drfuse}.

\begin{figure}
    \centering
    \begin{tabular}{ccc}
      \includegraphics[width=0.25\linewidth, valign=c]{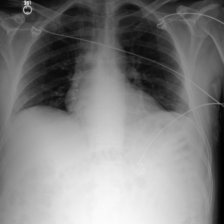}& \includegraphics[width=0.25\linewidth, valign=c]{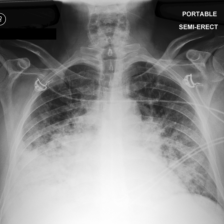} & \includegraphics[width=0.25\linewidth, valign=c]{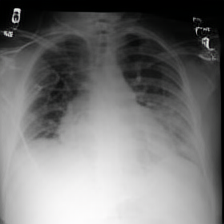}\\
        (a) Initial Chest X-ray & (b) CXR taken after 34 hours & (c) Generated by \ddlcxr 
    \end{tabular}
    \caption{A real ICU patient with rapid CXR changes. (a) \textit{Initial radiology findings}: Low lung volumes but lungs are clear of consolidation or pulmonary vascular congestion. No acute cardiopulmonary process. (b) \textit{Radiology findings after 34 hours}: Severe relatively symmetric \textbf{bilateral pulmonary consolidation}. (c) CXR generated by \ddlcxr given the initial CXR image shown in (a) and the EHR data within the 34 hours. Clear signs of bilateral pulmonary consolidation can be seen from the generated image. The visualization shows that 
    \textbf{\ddlcxr could generate updated CXR images that respect the anatomical structure of the patient and reflect the disease progression.}
    }\label{fig:motivating_example}
\end{figure} 

Despite the promising results obtained, the inherent asynchronicity of multimodal clinical data still hinders effective integration. Take the intensive care unit (ICU) setting as an example, patients are subject to continuous monitoring systems that capture vital signs, including heart rate, blood pressure, and oxygen saturation, with this information being routinely recorded in the EHR~\cite{adler2015electronic, choi2022advantage}. On the other hand, CXRs are captured only on an as-needed basis and often as less as possible, due to limitations of radiation dose and resources~\cite{henschke1996accuracy}. However, patients admitted to ICU are in life-threatening conditions, which means their medical status is prone to rapid changes and highly time-sensitive~\cite{zhang2023improving}. In the MIMIC-CXR dataset~\cite{johnson2019mimic}, it is observed that among patients with positive disease findings in their CXR, over 70\% of subsequent CXR images --- taken within a median interval of less than 24 hours --- exhibit changes in CXR findings. This implies that when a clinical prediction is needed, CXRs captured even only a few hours ago could have become outdated, especially for ICU patients who commonly have respiratory, cardiac, infectious, and traumatic conditions~\cite{ganapathy2012routine}. \cref{fig:motivating_example} shows such an example of a real patient in the MIMIC dataset.

\paragraph{Motivation} Existing works adopt the ``carry-forward" approach, i.e., using the last CXR image available for downstream prediction tasks~\cite{grant2021deep, mlhc2022hayatmedfuse}. This strategy ignores the potential rapid changes between the prediction time and the time of the last CXR image taken and thus inevitably leads to suboptimal prediction performance.
On the contrary, we hypothesize that generating an updated CXR image at the prediction time could mitigate the asynchronicity problem and enhance the prediction accuracy. Nevertheless, generating patient-specific CXR images presents unique challenges. While multimodal generation has been explored extensively in various fields, these methods are not readily adaptable for generating individualized CXR images. In domains such as text-to-audio~\cite{audioldm} or text-to-image generation~\cite{ldm}, the attributes that need to be controlled (e.g., painting style) can be explicitly defined in input modalities (e.g., the text prompt). However, in the clinical context, explicit descriptions of a patient's anatomical structures, organ functions, and disease progression, which are highly specific to individual patients and critical for downstream prediction, are not directly available. 

\paragraph{Contribution} To tackle the aforementioned challenge, we propose \textit{\underline{D}iffusion-based \underline{D}ynamic \underline{L}atent \underline{C}hest \underline{X}-\underline{r}ay Image Generation} (\ddlcxr)\footnote{The code is available at \url{https://github.com/Chenliu-svg/DDL-CXR}.}, which utilizes a tailored latent diffusion model (LDM)~\cite{ldm} to generate individualized CXR images for clinical prediction. Specifically, \ddlcxr learns to generate representations in a latent space encoded by a variational auto-encoder (VAE). To incorporate detailed information about the patient’s anatomical structure and organ specifics, we use a previous CXR image from the same patient as the reference image. To generate latent representations that align with the disease progression, we use a Transformer model~\cite{vaswani2017attention} to encode the irregular EHR data spanning from the reference CXR to the prediction time. To further capture the implicit interactions between EHR and CXR, we use the encoded EHR representation to predict the labels of abnormality finding of the target image. To force the LDM to capture the disease course in the EHR data, we explore a contrastive learning approach for training the LDM. The generated up-to-date latent CXR is later fused with historical data for downstream clinical prediction. 

We summarize our contributions as follows:
\begin{itemize}[leftmargin=*, itemsep=0em, parsep=0pt, topsep=0pt, partopsep=0pt]
    \item To our knowledge, \ddlcxr is the first work to generate an updated individual CXR image to improve clinical multimodal fusion, thereby alleviating the asynchronicity between EHR and CXR.
    \item We propose a contrastive learning approach for the LDM training to enable the disease course in EHR to be captured and utilized by the LDM.
    \item Experiments show that \ddlcxr outperforms existing methods in both multi-modal clinical prediction and individual CXR generation. 
\end{itemize}

\section{Related Work}
\paragraph{Clinical multi-modal fusion}
Integrating multi-modal clinical data has shown beneficial for various clinical prediction tasks~\cite{yang2023manydg}, including COVID-19 prediction~\cite{jiao2021prognostication}, pulmonary embolism diagnosis~\cite{huang2020multimodal,zhi2022multimodal}, AD diagnosis~\cite{daft-polsterl2021combining} and X-ray image abnormality detection~\cite{hsieh2023mdf}. 

Different strategies have been proposed to facilitate the fusion of multi-modal clinical data~\cite{EHRimagefusingreview,cui2023deep}. 
\citet{mlhc2022hayatmedfuse} adopts feature-level fusion with an LSTM layer, while \citet{zhang2022mmformer} utilizes a modality-correlated encoder to capture long-range dependencies across modalities. \citet{zhang2022m3care} and \citet{edward2023} incorporate modality type embedding into the self-attention to capture the interaction. Despite the effort, existing methods for multi-modal fusion are driven only by downstream predictions. How to capture the more fundamental interaction between different data modalities remains an open challenge.

In the temporal setting, asynchronicity presents another major challenge. Unlike the settings of medical images and radiology reports~\cite{BioViL-T}, which are naturally aligned in time, EHR and CXR are often highly asynchronous, bringing extra difficulties to information integration. ``Carry-forward'' is a common strategy adopted, where the last available data from different modalities are used~\cite{mlhc2022hayatmedfuse,yao2024drfuse}.  \citet{edward2023} and \citet{zhang2023improving} also adopt this approach while modeling the time information of the last available data.

\paragraph{Conditional latent diffusion models}
The diffusion model is one of the state-of-the-art generative models~\cite{ddpm, ddim} that has found important applications in areas such as image generation~\cite{saharia2022photorealistic}, sound generation~\cite{yang2023diffsound}, joint audio and video generation~\cite{ruan2023mm}, and tabular data generation~\cite{kotelnikov2023tabddpm}. 
To reduce the computational cost, LDM~\cite{ldm} proposes to train diffusion models on a latent space encoded via pre-trained VAE, thus improving training and sampling efficiency as well as preserving generation quality. It also incorporates an attention mechanism into its underlying neural backbone to allow more flexible conditioning. 

Based on LDM, multi-modal generation models have been developed using priors obtained from large-scale contrastive pre-training, e.g., contrastive-image pairs for text-to-image generation~\cite{ramesh2022hierarchical} and contrastive language-audio pairs for text-to-audio generation~\cite{audioldm}. However, it is infeasible to apply this method to clinical settings since many clinical data modalities, e.g., CXR and EHR, capture different aspects of patients and cannot be semantically aligned like the image and caption pairs as in CLIP.  

In clinical settings, LDM-based models are developed for brain MRI image generation, conditioned on age, sex, brain structure volumes~\cite{pinaya2022brain}, and a subset of MRI slices~\cite{peng2023generating}. 
For CXR image generation, \citet{packhauser2023generation} adopts a thoracic abnormality classifier-aided LDM to generate anonymous CXR images for privacy-protected data generation. \citet{weber2023cascaded} utilizes pathology labels, radiological reports, and radiologists' annotations for synthesizing customized CXR images. \citet{gu2023biomedjourney} explores counterfactual generation for CXR using information from imaging reports. Generating individual CXR images that reflect disease courses in EHR and applying them to medical predictions remains an open challenge.

\begin{figure}
    \centering
    \includegraphics[width=0.98\textwidth]{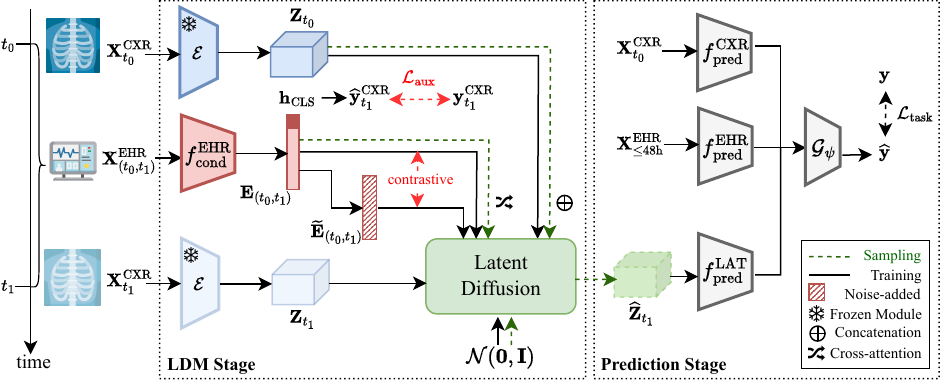}
    \caption{The overview of the proposed framework \ddlcxr. It consists of two stages. The \textbf{LDM stage} learns to generate an individualized up-to-date latent CXR at time $t_1$, $\hat{\mathbf{Z}}_{t_1}$, to address asynchronicity by conditioning on a previous CXR image taken at time $t_0$, $\mathbf{X}^{\text{CXR}}_{t_0}$, which provides the anatomical structure of the patient, as well as EHR data between $t_0$ and $t_1$, $\mathbf{X}^{\text{EHR}}_{(t_0, t_1)}$, that provides information on disease progression. A contrastive loss and auxiliary loss are enforced for better EHR information integration. The generation module encapsulates cross-modal interactions to assist in clinical prediction. The \textbf{prediction stage} fuses the generated latent CXR, the most recent CXR image, and the complete EHR time series for clinical predictions. }
    \label{fig:overview}
\end{figure}

\section{\ddlcxr: The Proposed Method}
\label{sec:method}
In this work, we focus on improving multimodal clinical predictions by generating latent CXR images that are in line with patient conditions at prediction time. The generation process also works as a fusion module that captures the cross-modal interaction between EHR and CXR. The overview of \ddlcxr is depicted in \cref{fig:overview}. It consists of two stages: the LDM stage and the prediction stage. In the LDM stage, we use the consecutive image pairs to train an LDM that generates representations within a latent space encoded by a variational autoencoder (VAE). To generate patient-specific CXRs, an earlier CXR image of the same patient is used as a reference to capture the anatomical structure, and the EHR time series between the consecutive image pairs is used to capture the disease progression. In the prediction stage, conditioned on this composite information, \ddlcxr generates updated and informative CXR representations at prediction time, which are subsequently fused with available EHR data as well as the previous CXR image for downstream prediction tasks.

\subsection{Notations and Preliminaries} \label{sec:method-notations}
\paragraph{EHR and CXR data} Patient-wisely, we denote the EHR time series within the time interval between $t_i$ and  $t_j$ as $\mathbf{X}^{\text{EHR}}_{(t_i,t_j)}=[\mathbf{x}_{t_i},\mathbf{x}_{t_{i}+1},\dots,\mathbf{x}_{t_j} ]$, where $\mathbf{x}_t\in\mathbb{R}^K$ is the variables recorded at time $t$ and $K$ is the number of features. 
We denote the grayscale CXR images taken at the time $t_i$ as $\mathbf{X}^{\text{CXR}}_{t_i} \in \mathbb{R}^{W\times H}$, where $W$ and $H$ denote its width and height, respectively. For each CXR image, we extract the abnormality finding label, $\mathbf{y}^{\text{CXR}}_{t_i}$,  from radiology reports using CheXpert~\cite{irvin2019chexpert}.
\paragraph{Predictive latent space for CXR} 
Using diffusion models in a semantic latent space, rather than a high-dimensional data space, has shown a substantial decrease in computational expenses with minimal impact on synthesis quality~\cite{ldm}. To obtain an informative and expressive latent space, we first train a VAE \cite{vae} consisting of an encoder $\mathcal{E}$ and a decoder $\mathcal{D}$. 
The data used for training VAE are all available CXR images in the training set with corresponding abnormality finding labels, $\left(\mathbf{X}^{\text{CXR}}_t, \mathbf{y}^{\text{CXR}}_t \right)$. 
The primary objective of the VAE is to reconstruct the original CXR image $\mathbf{X}^{\text{CXR}}_t$ with $\mathcal{D}(\mathcal{E}(\mathbf{X}^{\text{CXR}}_t))$. We follow the VAE training process in \cite{ldm}, incorporating a pixel-wise reconstruction loss accompanied by a perceptual loss~\cite{Zhang_2018_CVPR}, an adversarial objective, and a lightly-penalized Kullback-Leibler loss towards a standard normal aiming at constraining the latent spaces from excessively high variance.
Besides, to improve the encoder's ability to predict, we also include a prediction loss regarding the abnormality label $\mathbf{y}^{\text{CXR}}_t$. 
We denote the encoded latent CXR by $\mathbf{Z}_t=\mathcal{E}(\mathbf{X}^{\text{CXR}}_t) \in \mathbb{R}^{C\times \frac{W}{r} \times \frac{H}{r}}$, where $C$ represents the channel of the compressed representation and $r$ represents the compression ratio. We first pre-train the VAE model and then freeze it throughout the training and inference of \ddlcxr. More details on VAE training are presented in \cref{app:A1-training}.

\subsection{LDM Stage: Dynamic Latent CXR Generation}

As discussed previously, to generate an up-to-date, patient-specific latent CXR, it is important to incorporate the unique anatomical details of the individual patient. Furthermore, the generated image must accurately reflect the evolving pathology as documented in the irregular EHR time series. To this end, we extract all sequential image pairs and the EHR time series between them. We denote each sample as a quadruplet: $\big(\mathbf{X}^{\text{CXR}}_{t_0}, \mathbf{X}^{\text{EHR}}_{(t_0, t_1)}, \mathbf{X}^{\text{CXR}}_{t_1}, \mathbf{y}^{\text{CXR}}_{t_1} \big)$. CXR images are encoded using the pre-trained VAE as we aim to generate latent CXR images: $
    \mathbf{Z}_{t_0} = \mathcal{E}(\mathbf{X}^{\text{CXR}}_{t_0}), \mathbf{Z}_{t_1} = \mathcal{E}(\mathbf{X}^{\text{CXR}}_{t_1})
$.
We follow prior works on diffusion models to learn our LDM~\cite{ldm,ddpm}. 
It comes down to learning a network that predicts the noise added to the noisy latent $\mathbf{Z}_{t_1}^{(n)}$ at denoising step $n$ as
$\boldsymbol{\epsilon}_\theta\left(\mathbf{Z}_{t_1}^{(n)},\mathbf{Z}_{t_0},f^{\text{EHR}}_{\text{cond}}(\mathbf{X}_{(t_0,t_1)}^{\text{EHR}}), n\right)$. {Following prior works~\cite{ldm,ddpm}, we parameterize $\boldsymbol{\epsilon}_\theta$ by a standard UNet~\cite{ronneberger2015u}.} Here $f^{\text{EHR}}_{\text{cond}}(\cdot)$ is the encoder for the irregular EHR time series to be detailed later. The detailed diffusion and denoising processes are presented in \cref{app:A1-training}.

\paragraph{Neural backbone and conditioning mechanisms}
Due to the remarkable capability of UNet~\cite{ronneberger2015u} in capturing the spatial structure of images, we follow prior works and use a UNet as our neural backbone $\boldsymbol{\epsilon}_\theta$.
It predicts the noise added in the diffusion process, conditioned on the reference image and the EHR time series. To explicitly capture and utilize the anatomical structure of individual patients, we first concatenate the reference latent CXR $\mathbf{Z}_{t_0}$  and the step-$n$ noisy latent $\mathbf{Z}_{t_1}^{(n)}$. To further integrate the disease course embedded in the EHR time series, we use the cross-attention mechanism to capture the interaction between the two modalities. Formally, the input to the UNet layers is given by:
\begin{equation}\small
    \begin{aligned}
\operatorname{Attention}(\mathbf{Q}, \mathbf{K}, \mathbf{V}) &= \operatorname{softmax}\Big(\frac{\mathbf{Q}\mathbf{K}^\top}{\sqrt{d}}\Big) \cdot \mathbf{V}, \\ 
\text{with }\mathbf{Q} = \mathbf{W}_Q \cdot \varphi \left(\mathbf{Z}_{t_1}^{(n)}||\mathbf{Z}_{t_0}\right),
\mathbf{K} &= \mathbf{W}_K \cdot f^{\text{EHR}}_{\text{cond}}(\mathbf{X}_{(t_0,t_1)}^{\text{EHR}}),
\mathbf{V} = \mathbf{W}_V \cdot f^{\text{EHR}}_{\text{cond}}(\mathbf{X}_{(t_0,t_1)}^{\text{EHR}}), \\
\end{aligned}
\end{equation}
where $\varphi(\cdot)$ denotes the flattened intermediate representation of the UNet and $||$ denotes concatenation.

\paragraph{Capturing disease course via EHR time series} \label{sec:ehr_encoder}  To effectively capture useful information on disease progression for future CXR generation, we adopt a multi-task Transformer-based time series encoder~\cite{zerveas2021transformer} with the masked self-attention mechanism to handle the variable length of EHR time series~\cite{transformer}.
The encoded representation of EHR, $\mathbf{E}_{(t_0,t_1)}$, is given by
\begin{equation}\label{eqn:EHR_encoder}\small
 \mathbf{E}_{(t_0,t_1)} = f^{\text{EHR}}_{\text{cond}}(\mathbf{X}_{(t_0,t_1)}^{\text{EHR}}) = \text{Transformer}\left([\mathbf{h}_\text{CLS},\phi(\mathbf{x}_{t_0}), \dots,\phi(\mathbf{x}_{t_1})] \right),
\end{equation}
where $\phi(\mathbf{x}_{t})$ projects the original EHR time series into an embedding space and applies the positional encoding at time step $t$. $\mathbf{h}_\text{CLS}$ is the class token. 

To further extract information that is relevant to CXR generation and facilitate modality fusion at the LDM stage, we incorporate an auxiliary prediction task: using the class token from the encoded EHR to predict the abnormality findings $\mathbf{y}^{\text{CXR}}_{t_1}$, associated with the CXR image $\mathbf{X}^{\text{CXR}}_{t_1}$, i.e., $\widehat{\mathbf{y}}^{\text{CXR}}_{t_1}=g(\mathbf{h}_\text{CLS})$, where $g$ denotes the prediction function, e.g., an MLP, which is trained by jointly minimize the loss function given by
 $\mathcal{L}_{\text{aux}}:=\frac{1}{M}\frac{1}{L}\sum_{m=1}^{M}  \sum_{l=1}^{L} y^{\text{CXR}}_{ml}\log(\widehat{y}^{\text{CXR}}_{ml})+(1-y^{\text{CXR}}_{ml})\log(1- \widehat{y}^{\text{CXR}}_{ml}),$
where $M$ is the number of training samples for LDM and $L$ is the number of classes of abnormality labels of CXR.
The auxiliary task enables the EHR encoder to extract CXR-related information, which further encourages the interaction between EHR and CXR to be captured in the subsequent generation. 

\paragraph{Enhancing semantic multimodal fusion via contrastive LDM learning} The generation conditioning on EHR data is challenging because the EHR and CXR data are highly heterogeneous and the interactions are implicit. To force the LDM to utilize EHR information during generation, we propose a contrastive way of learning the conditional LDM. Specifically, for each EHR time series, we obtain a perturbed version of its representation $\widetilde{\mathbf{E}}_{(t_0,t_1)}=(1-\beta)\mathbf{E}_{(t_0,t_1)}+\beta\boldsymbol{\delta}$, where $\boldsymbol{\delta}\sim\mathcal{N}(\mathbf{0}, \mathbf{I})$ is randomly drawn from a standard normal distribution, $\beta$ is a hyperparameter controlling the strength of the noise. When the perturbed EHR is given as input, we expect the generated image to be far away from the target image. This leads to the following training objective function:
\begin{equation}
\resizebox{0.94\linewidth}{!}{$
    \begin{aligned}
\mathcal{L}_{\text{LDM}}:=&\mathbb{E}_{\mathbf{Z}_{t_1},\mathbf{Z}_{t_0},\mathbf{X}^{\text{EHR}}_{(t_0,t_1)},\boldsymbol{\epsilon}\sim\mathcal{N}(\mathbf{0},\mathbf{I}),n}
    \Bigg[  \left\|\boldsymbol{\epsilon}-\boldsymbol{\epsilon}_\theta\left(\mathbf{Z}_{t_1}^{(n)},\mathbf{Z}_{t_0},f^{\text{EHR}}_{\text{cond}}(\mathbf{X}^{\text{EHR}}_{(t_0,t_1)}), n \right) \right\|_2^2 \\
    & + \lambda_1 \max \left(  \left\|\boldsymbol{\epsilon} - \boldsymbol{\epsilon}_\theta\left(\mathbf{Z}_{t_1}^{(n)},\mathbf{Z}_{t_0},\mathbf{E}_{(t_0,t_1)},n \right) \right\|_2^2  - \left\|\boldsymbol{\epsilon} - \boldsymbol{\epsilon}_\theta\left(\mathbf{Z}_{t_1}^{(n)},\mathbf{Z}_{t_0},\widetilde{\mathbf{E}}_{(t_0,t_1)},n \right) \right\|_2^2+\alpha, 0\right) \Bigg],
    \end{aligned}
$}
    \label{eqn:ldm}
\end{equation}
where $\alpha$ is a hyperparameter controlling the tolerance of the noisy-conditional generation. $\lambda_1$ is a coefficient controlling the strength of the contrastive term. To ensure stability during training, we set the initial value of $\lambda_1$ to zero and linearly increase it to one during training. 

\subsection{Prediction Stage}
In the prediction stage, we do not have access to an up-to-date CXR image. Therefore, we generate an updated latent CXR $\widehat{\mathbf{Z}}_{t_1}$ at the prediction time $t_1$ using the last available CXR image $\mathbf{X}^{\text{CXR}}_{t_0}$ as the reference image and the EHR time series in between $\mathbf{X}^{\text{EHR}}_{(t_0, t_1)}$. To make predictions using available EHR, we adopt another time series encoder, $f^{\text{EHR}}_{\text{pred}}$, which has the same structure as $f^{\text{EHR}}_{\text{cond}}$ as in ~\cref{eqn:EHR_encoder}. Note that for prediction, the EHR data used, $\mathbf{X}^{\text{EHR}}_{\leq \text{48h}}$ covers all EHR time series with the observation time set as 48 hours,  ensuring the available information is fully utilized. In other words, $\mathbf{X}^{\text{EHR}}_{(t_0, t_1)}\subseteq \mathbf{X}^{\text{EHR}}_{\leq \text{48h}}$. 
In clinical practice, clinicians make predictions not only based on the latest CXR, but also on past CXR images as reference for disease basis. To this end, we employ all available data: $\mathbf{X}^{\text{CXR}}_{t_0}$, $\mathbf{X}^{\text{EHR}}_{\leq \text{48h}}$ and the generated latent CXR $\widehat{\mathbf{Z}}_{t_1}$ to make the final clinical prediction: 
\begin{equation}\small
\widehat{\mathbf{y}}=\mathcal{G}_{\psi}\left(f^{\text{CXR}}_{\text{pred}}(\mathbf{X}^{\text{CXR}}_{t_0}), \ f^{\text{EHR}}_{\text{pred}}(\mathbf{X}^{\text{EHR}}_{\leq \text{48h}}), \ f^{\text{LAT}}_{\text{pred}}(\widehat{\mathbf{Z}}_{t_1})\right).
\end{equation}
Here $ f^{i}_{\text{pred}}$, $i\in \left\{ \text{CXR, EHR, LAT}\right\} $ are encoders for CXR, EHR, and the generated latent CXR, accordingly. We parameterize $f^{\text{LAT}}_{\text{pred}}$ and $f^{\text{EHR}}_{\text{pred}}$ using Transformer models, and $f^{\text{CXR}}_{\text{pred}}$ using a ResNet model. The predicting model $\mathcal{G}_\psi$ with $\psi$ denoting the model parameter, is parameterized by a self-attention layer. We learn it by minimizing the cross-entropy (CE) loss:
\begin{equation}\small
    \mathcal{L}_{\text{task}}:=\sum_{m=1}^{M^\prime}  \sum_{l=1}^{L^\prime} y_{ml} \log(\widehat{y}_{ml})+(1-y_{ml})\log(1- \widehat{y}_{ml}),
    \label{eq:ldm_loss}
\end{equation}
where $L^\prime$ is the number of classes in the prediction task and $M^\prime$ is the number of training samples in the prediction stage.

\section{Experiments}
\label{sec:exp}
\subsection{Experiment Settings}
\label{sec:exp-set}
\paragraph{Datasets}
We empirically evaluate the clinical predictive performance of \ddlcxr using MIMIC-IV \cite{johnson2023mimic} and MIMIC-CXR \cite{johnson2019mimic}\footnote{Both are open source under the PhysioNet Credentialed Health Data License 1.5.0 license.}. MIMIC-IV comprises de-identified critical care data from adult patients admitted to either ICUs or the emergency department (EDs) of Beth Israel Deaconess Medical Center (BIDMC) between 2008 and 2019, and MIMIC-CXR contains chest X-rays and reports collected from BIDMC, with a subset of patients matched with those in MIMIC-IV. 
For EHR data, we follow a preprocessing pipeline similar to that described in \cite{mlhc2022hayatmedfuse}. 17 clinical time series variables as well as age and gender are extracted. The details can be found in \cref{app:A2-data}.

\paragraph{Dataset construction and partition}
The inclusion criteria for this study involve ICU stays from the matched subset of MIMIC-IV and MIMIC-CXR that contain at least one CXR image (with Anterior-Posterior (AP) projection) during the ICU stay or within 24 hours before ICU admission. We exclude ICU stays with lengths shorter than 48 hours.
The dataset is randomly split by the patient identifier with a ratio of 24:4:7 for training, validation, and testing, which avoids patient overlapping between subsets.

From the training patients, we further extract data for training the VAE, the LDM, and the prediction model. We extract all images from the training patients for training VAE and extract all CXR image pairs of the same patient taken at any interval greater than 12 hours for training the LDM, i.e.,
\begin{equation*}\small
    \mathcal{D}_{\text{LDM}}=\left\{\left(\mathbf{X}^{\text{CXR}}_{t_0},\mathbf{X}^{\text{CXR}}_{t_1}, \mathbf{X}^{\text{EHR}}_{(t_0,t_1)}, \mathbf{y}^{\text{CXR}}_{t_1} \right)_{(t_1 - t_0 )> \text{12h}}\right\},
\end{equation*}
where a single ICU stay may contain multiple data pairs for LDM training. This greatly enlarges the training subset for the LDM stage. 

For the prediction stage, we extract the last available CXR image and the EHR time series in the first 48 hours and the label for the prediction task of each ICU stay, i.e., the triplet $\left( \mathbf{X}^{\text{CXR}}_{\text{last}}, \mathbf{X}^{\text{EHR}}_{\leq\text{48h}}, \mathbf{y}_{\text{task}} \right)$. Note that the EHR time series used in the prediction stage differs from that in the LDM stage in their time interval since they serve for different purposes. 

We use the same approach to extract the validation subsets for hyperparameter tuning of VAE, LDM, and the prediction model. Note that the testing patients are held out for evaluating prediction performance only, and are not involved in the training and model selection of VAE and LDM.

\paragraph{Prediction tasks and evaluation metrics}
We evaluate \ddlcxr with two clinical prediction tasks: in-hospital mortality prediction and phenotype classification using clinical data collected within the first 48 hours of ICU admissions. 
The phenotype classification is a multi-label classification task, where the labels are defined by the 25 disease phenotypes, extracted following~\cite{harutyunyan2019multitask}. 
The details of the label prevalence and data cohort statistics can be found in \cref{app:A2-data}.

We evaluate the performance using two metrics, the Area Under the Precision-Recall Curve (AUPRC) and the Area Under the Receiver Operating Characteristics (AUROC). For the phenotyping task, we report macro-averaged scores. We conduct each prediction experiment five times with distinct random seeds and reported the mean and standard deviation of the results. 

\paragraph{Baseline Models}
We compare the following methods. (1) \textbf{Uni-EHR}, a single-modal classifier for EHR time series based on Transformer~\cite{vaswani2017attention}, (2) \textbf{MMTM}~\cite{joze2020mmtm}, a multi-modal fusion method based on CNNs through squeeze and excitation operations, (3) \textbf{DAFT}~\cite{daft-polsterl2021combining} a general-purpose module for fusing tabular clinical information and image data by dynamically rescaling and shifting the feature maps of a convolutional layer, (4) \textbf{MedFuse}~\cite{mlhc2022hayatmedfuse}, an LSTM-based multimodal fusion method developed for clinical prediction using EHR and CXR, (5) \textbf{DrFuse}~\cite{yao2024drfuse}, a disentangled learning approach that handles modality missing and modal inconsistency in clinical multi-modal fusion, and (6) \textbf{GAN-based generation}~\cite{xia2021learning}, a model originally proposed to generate individual brain images conditioning on age and Alzheimer’s Disease (AD) status via training a conditional GAN.

\subsection{Prediction Performance}

\begin{table}\small
    \centering
    \caption{Overall performance for the phenotype classification and mortality prediction task as measured by AUPRC and AUROC scores. \ddlcxr outperforms all baselines in these metrics.}
    \label{tab:overall}
    \begin{tabular}{ccccc}
    \toprule
     &  \multicolumn{2}{c}{Phenotyping} &  \multicolumn{2}{c}{Mortality} \\\midrule
               & AUPRC                                 & AUROC     & AUPRC                                 & AUROC                             \\\midrule
Uni-EHR~\cite{vaswani2017attention} & 0.434 {\small$\pm$0.009} & 0.720 {\small$\pm$0.006} & 0.498 {\small$\pm$0.007} & 0.815 {\small$\pm$0.007}\\
MMTM~\cite{joze2020mmtm} & 0.430 {\small$\pm$0.005} & 0.715 {\small$\pm$0.003} & 0.422 {\small$\pm$0.014} & 0.785 {\small$\pm$0.004}\\
DAFT~\cite{daft-polsterl2021combining} & 0.435 {\small$\pm$0.002} & 0.720 {\small$\pm$0.003} & 0.448 {\small$\pm$0.004} & 0.800 {\small$\pm$0.003}\\
MedFuse~\cite{mlhc2022hayatmedfuse} & 0.437 {\small$\pm$0.001} & 0.718 {\small$\pm$0.002} & 0.443 {\small$\pm$0.009} & 0.793 {\small$\pm$0.003}\\
DrFuse~\cite{yao2024drfuse} & 0.459 {\small$\pm$0.003} & 0.729 {\small$\pm$0.004} & 0.460 {\small$\pm$0.004} & 0.773 {\small$\pm$0.008}\\
GAN-based~\cite{xia2021learning} & 0.453 {\small$\pm$0.010} & 0.728 {\small$\pm$0.008} & 0.505 {\small$\pm$0.018} & 0.816 {\small$\pm$0.010}\\
\ddlcxr (ours) & \textbf{0.470} {\small$\pm$0.003} & \textbf{0.740} {\small$\pm$0.002} & \textbf{0.523} {\small$\pm$0.011} & \textbf{0.822} {\small$\pm$0.009}\\\bottomrule
    \end{tabular}
\end{table}

\paragraph{\ddlcxr obtains the best overall performance.} We summarize the overall performance of the phenotype classification and in-hospital mortality prediction in \cref{tab:overall}, where \ddlcxr outperforms all baselines. This shows that generating an updated CXR during test time is beneficial for downstream tasks. On the contrary, DrFuse, MedFuse, DAFT, and MMTM use the last available CXR for prediction, which might have been outdated. 

The performance gain of \ddlcxr in terms of AUPRC is particularly noteworthy as the AUPRC metric is especially relevant in the context as it underscores the effectiveness of our approach in identifying the positive class in imbalanced medical datasets. \ddlcxr achieves relative improvements of 2.4\% and 3.56\% over the best baselines in terms of AUPRC for phenotype classification and mortality prediction, respectively.

\begin{table}\small
    \centering
    \caption{The mean of AUROC score with standard deviation for mortality prediction for overall and different time gaps. $\delta$ represents the time interval (by hour) between the prediction time and the time of the last available CXR. Numbers in bold indicate the best performance in each column. \ddlcxr outperforms all baselines in most settings. The AUPRC scores can be found in \cref{app:B1-AUPRC}.}\label{tab:mortality_auroc}
    \resizebox{0.99\linewidth}{!}{
    \begin{tabular}{cccccc}
    \toprule
  & Overall & $\delta<12$ & $12 \leq\delta <24$ & $24 \leq\delta <36$ & $\delta \geq 36$\\ 
\textit{prevalence} & 14.7\% & 16.6\% & 19\% & 15.9\% & 9.26\%\\ \midrule
Uni-EHR~\cite{vaswani2017attention} & 0.815 {\small$\pm$0.007} & 0.854 {\small$\pm$0.010} & 0.799 {\small$\pm$0.013} & 0.756 {\small$\pm$0.019} & 0.796 {\small$\pm$0.008}\\
MMTM~\cite{joze2020mmtm} & 0.785 {\small$\pm$0.004} & 0.798 {\small$\pm$0.008} & 0.763 {\small$\pm$0.004} & 0.760 {\small$\pm$0.012} & 0.772 {\small$\pm$0.014}\\
DAFT~\cite{daft-polsterl2021combining} & 0.800 {\small$\pm$0.003} & 0.803 {\small$\pm$0.010} & 0.782 {\small$\pm$0.009} & \textbf{0.776} {\small$\pm$0.006} & 0.796 {\small$\pm$0.008}\\
MedFuse~\cite{mlhc2022hayatmedfuse} & 0.793 {\small$\pm$0.003} & 0.812 {\small$\pm$0.004} & 0.762 {\small$\pm$0.007} & 0.760 {\small$\pm$0.009} & 0.800 {\small$\pm$0.010}\\
DrFuse~\cite{yao2024drfuse} & 0.773 {\small$\pm$0.008} & 0.802 {\small$\pm$0.012} & 0.717 {\small$\pm$0.023} & 0.757 {\small$\pm$0.041} & 0.723 {\small$\pm$0.013}\\
GAN-based~\cite{xia2021learning} & 0.816 {\small$\pm$0.010} & 0.846 {\small$\pm$0.010} & \textbf{0.800} {\small$\pm$0.011} & 0.760 {\small$\pm$0.026} & 0.806 {\small$\pm$0.016}\\\midrule
\ddlcxr (ours) & \textbf{0.822} {\small$\pm$0.009} & \textbf{0.867} {\small$\pm$0.015} & \textbf{0.800} {\small$\pm$0.008} & 0.753 {\small$\pm$0.015} & \textbf{0.830} {\small$\pm$0.011}\\\bottomrule
    \end{tabular}
    }
\end{table}

\paragraph{Mortality prediction with varying time interval}
We define the time interval (by hour) between the prediction time and the time of the last available CXR as $\delta$ and compute the evaluation metrics in patient groups with different ranges of $\delta$. The results are presented in \cref{tab:mortality_auroc}. Since the label prevalence varies significantly between groups, making the comparison of AUPRC between groups less meaningful, we report AUROC in the paper and AUPRC in the appendix. \ddlcxr consistently outperforms the baseline models for most groups of $\delta$ for the mortality prediction task. 
As the $\delta$ increases, the last CXR becomes more ``outdated'', and we observe a noticeable increase in the performance gap between the best baseline and \ddlcxr in the group ($\delta \geq 36$), from 0.806 to 0.830. This validates our hypothesis that the generation of a timely CXR, accounting for disease progression, can significantly enhance the performance of clinical predictions.

\begin{table}\small
\setlength{\tabcolsep}{3pt}
    \centering
    \caption{The AUPRC score of predicting each phenotype label. \ddlcxr obtains the highest average rank. Full names of phenotype labels and AUROC scores can be found in the Appendix.}
    \label{tab:disease_prauc}
    \resizebox{\linewidth}{!}{
    \begin{tabular}{lccccccc}\toprule
    
 & Uni-EHR & MMTM & DAFT & MedFuse & DrFuse & GAN-based & \ddlcxr\\\midrule
Acute renal failure & \textit{0.573} & 0.568 & 0.572 & 0.565 & 0.564 & 0.563 & \textbf{0.588}\\
Acute cerebrovascular disease & 0.425 & 0.418 & 0.419 & \textit{0.434} & 0.399 & \textbf{0.446} & 0.416\\
Acute myocardial infarction & 0.185 & 0.192 & 0.187 & \textbf{0.219} & \textit{0.209} & 0.171 & 0.206\\
Cardiac dysrhythmias & 0.579 & 0.532 & 0.548 & 0.560 & \textit{0.584} & 0.561 & \textbf{0.605}\\
Chronic kidney disease & 0.515 & 0.505 & \textit{0.515} & 0.497 & 0.477 & 0.501 & \textbf{0.538}\\
COPD and bronchiectasis & 0.319 & 0.327 & 0.342 & 0.344 & \textbf{0.405} & 0.372 & \textit{0.382}\\
Surgical complications & 0.370 & 0.379 & \textit{0.385} & 0.381 & 0.377 & 0.344 & \textbf{0.388}\\
Conduction disorders & 0.276 & 0.287 & 0.298 & 0.286 & \textit{0.632} & 0.609 & \textbf{0.633}\\
CHF; nonhypertensive & 0.593 & 0.619 & 0.647 & 0.631 & \textit{0.661} & 0.652 & \textbf{0.682}\\
CAD & 0.560 & 0.540 & 0.556 & 0.544 & 0.581 & \textit{0.590} & \textbf{0.611}\\
DM with complications & \textit{0.562} & \textbf{0.569} & 0.552 & 0.561 & 0.550 & 0.552 & 0.524\\
DM without complication & \textbf{0.370} & 0.367 & 0.343 & 0.356 & \textit{0.369} & 0.352 & 0.368\\
Disorders of lipid metabolism & \textit{0.594} & 0.576 & 0.570 & 0.566 & 0.584 & 0.587 & \textbf{0.601}\\
Essential hypertension & 0.551 & 0.519 & 0.525 & 0.518 & 0.502 & \textit{0.554} & \textbf{0.561}\\
Fluid and electrolyte disorders & 0.655 & \textit{0.664} & 0.662 & 0.656 & 0.658 & 0.662 & \textbf{0.672}\\
Gastrointestinal hemorrhage & 0.180 & 0.142 & 0.162 & \textbf{0.192} & \textit{0.191} & 0.151 & 0.180\\
Secondary hypertension & \textit{0.463} & 0.455 & 0.452 & 0.453 & 0.437 & 0.451 & \textbf{0.484}\\
Other liver diseases & 0.316 & 0.316 & 0.341 & 0.344 & \textit{0.372} & 0.362 & \textbf{0.378}\\
Other lower respiratory disease & 0.219 & 0.209 & 0.206 & 0.223 & \textbf{0.255} & 0.236 & \textit{0.242}\\
Other upper respiratory disease & 0.166 & 0.137 & 0.166 & 0.202 & \textbf{0.274} & 0.196 & \textit{0.234}\\
Pleurisy; pneumothorax & 0.143 & 0.145 & 0.159 & 0.159 & \textbf{0.172} & \textit{0.171} & 0.166\\
Pneumonia & 0.412 & \textbf{0.437} & \textit{0.429} & 0.419 & 0.406 & 0.415 & 0.428\\
Respiratory failure & 0.655 & \textit{0.686} & 0.674 & 0.671 & \textbf{0.692} & 0.663 & 0.669\\
Septicemia (except in labor) & \textit{0.585} & 0.573 & 0.580 & 0.565 & 0.562 & 0.573 & \textbf{0.603}\\
Shock & \textit{0.590} & 0.584 & 0.582 & \textbf{0.592} & 0.572 & 0.587 & 0.586\\\midrule
\multicolumn{1}{r}{Average Rank} & 4.4 & 4.64 & 4.4 & 4.24 & \textit{3.88} & 4.16 & \textbf{2.28}\\\bottomrule
\end{tabular}}
\end{table}

\paragraph{Phenotype classification}
The class-wise AUPRC scores for the phenotyping task are detailed in Table \ref{tab:disease_prauc}, where \ddlcxr demonstrates notable performance improvements, achieving the highest average rank across all phenotype labels. Due to space limit, we report the standard deviations and the AUROC scores in \cref{app:B13-disease}. The improvement over baseline multimodal fusion methods validates the effectiveness of facilitating fusion between EHR and CXR in the presence of asynchronicity.

\subsection{Quality of Generated Chest X-ray Images} 
\paragraph{Quantitative Evaluation} We evaluate the quality of generated CXRs using the test set of the LDM stage, where the ground-truth target CXR is available. The Fréchet Inception Distance (FID) score~\cite{fid} evaluates the similarity between the distributions of generated and ground-truth target CXRs by computing Fréchet distance on the representation obtained from a pre-trained Inception-v3 network. 
\begin{wraptable}{r}{0.4\textwidth}\small
    \centering
    \setlength{\tabcolsep}{2pt}
    \caption{Generation quality.}
    \label{tab:gen_fid}
    \resizebox{0.4\textwidth}{!}{
    \begin{tabular}{lccc}\toprule
                   &  FID (\textdownarrow)   &FD (\textdownarrow)  &  WD (\textdownarrow)\\\midrule
        Last-CXR   &  16.50 & 2322.68 & 6353.06 \\\hdashline
        w/o $\mathbf{Z}_{t_0}$   & 47.91 & 3260.17 & 7491.17 \\
        w/o $\mathbf{E}_{(t_0, t_1)}$   &  \textbf{30.03}& 2412.80 & 7226.38 \\    
        GAN-based  &  98.67 & 3651.27 & 7922.71 \\\hdashline   
        \ddlcxr  &  {33.83} & \textbf{2316.08} & \textbf{7132.82} \\\bottomrule
    \end{tabular}}
\end{wraptable}
Besides, we directly measure the Fréchet distance (FD) and Wasserstein distance (WD) in the latent space of the VAE between the generated and the target CXRs. Results are shown in \cref{tab:gen_fid}. Results of ``Last-CXR'' are obtained between reference images $\mathbf{X}_0$ and target images $\mathbf{X}_1$, both are directly from the dataset without generation. Thus, this provides a reference to the lower bound of the metrics used. \ddlcxr surpasses GAN-based methods across all metrics and obtains the lowest FD and WD. ``w/o $\mathbf{Z}_{t_0}$'' and ``w/o $\mathbf{E}_{(t_0, t_1)}$'' are obtained by removing the condition of the last available CXR and EHR data, respectively. Notably, excluding EHR data from the generation conditions resulted in lower FID scores, which is natural since the generation becomes less restrictive.

\begin{figure}
    \centering
    \setlength{\tabcolsep}{1pt}
    \begin{tabular}{c c c c c c c c}
         & \textit{\small Sample \#1} & \textit{\small Sample \#2} & \textit{\small Sample \#3} & \textit{\small Sample \#4} & \textit{\small Sample \#5} & \textit{\small Sample \#6} & \textit{\small Sample \#7} \\
        {\small $\mathbf{X}^{\text{CXR}}_{t_0}$}
            & \includegraphics[width=0.125\linewidth, valign=c]{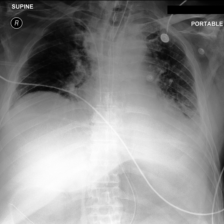} 
            & \includegraphics[width=0.125\linewidth, valign=c]{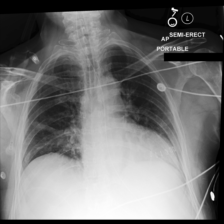} 
            & \includegraphics[width=0.125\linewidth, valign=c]{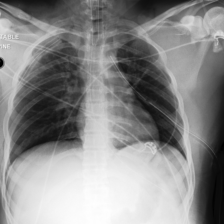}
            & \includegraphics[width=0.125\linewidth, valign=c]{motivation/1_x0_11-34.png}
            & \includegraphics[width=0.125\linewidth, valign=c]{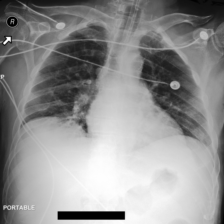}
            & \includegraphics[width=0.125\linewidth, valign=c]{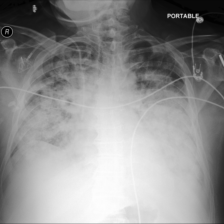}
            & \includegraphics[width=0.125\linewidth, valign=c]{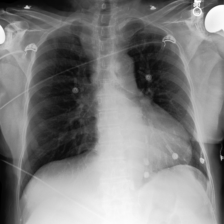}\\[-10pt]~\\
        {\small $\mathbf{X}^{\text{CXR}}_{t_1}$} 
            & \includegraphics[width=0.125\linewidth, valign=c]{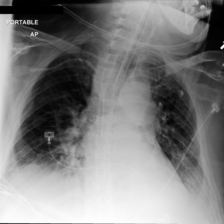} 
            & \includegraphics[width=0.125\linewidth, valign=c]{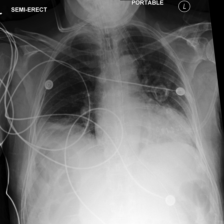} 
            & \includegraphics[width=0.125\linewidth, valign=c]{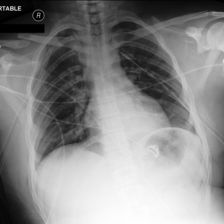}
            & \includegraphics[width=0.125\linewidth, valign=c]{motivation/1_x1_11-34.png}
            & \includegraphics[width=0.125\linewidth, valign=c]{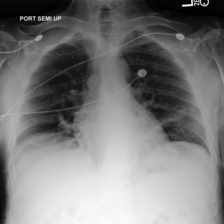}
            & \includegraphics[width=0.125\linewidth, valign=c]{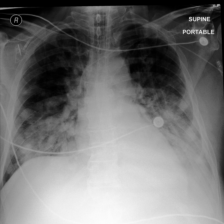}
            & \includegraphics[width=0.125\linewidth, valign=c]{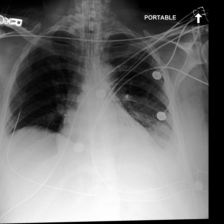}\\[-10pt]~\\
        {\small $\widehat{\mathbf{X}}^{\text{CXR}}_{t_1}$ }
            & \includegraphics[width=0.125\linewidth, valign=c]{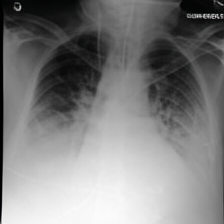} 
            & \includegraphics[width=0.125\linewidth, valign=c]{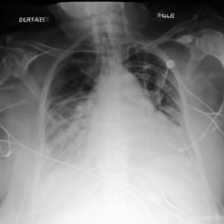} 
            & \includegraphics[width=0.125\linewidth, valign=c]{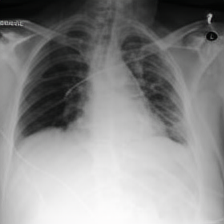}
            & \includegraphics[width=0.125\linewidth, valign=c]{motivation/1_gen_11-34.png}
            & \includegraphics[width=0.125\linewidth, valign=c]{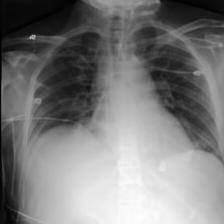}
            & \includegraphics[width=0.125\linewidth, valign=c]{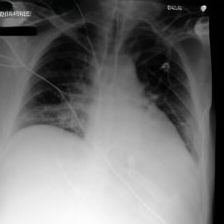}
            & \includegraphics[width=0.125\linewidth, valign=c]{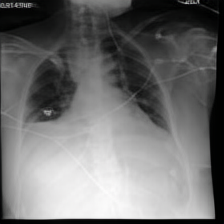}
    \end{tabular}
    \caption{Examples of images generated by \ddlcxr. From top to bottom, the three rows are reference images $\mathbf{X}^{\text{CXR}}_{t_0}$, ground-truth images $\mathbf{X}^{\text{CXR}}_{t_1}$, and generated images $\widehat{\mathbf{X}}^{\text{CXR}}_{t_1}$, respectively. The generations show that \ddlcxr captures the anatomical information from $\mathbf{X}^{\text{CXR}}_{t_0}$ and the information of disease progression extracted from EHR is blended well towards generating $\mathbf{X}^{\text{CXR}}_{t_1}$.}
    \label{fig:visual_examples}\vspace{-0.5em}
\end{figure}

\paragraph{Qualitative Evaluation} To further visually examine the generated CXR images, we decode the latent CXR and visualize seven examples in 
\cref{fig:visual_examples}. The first row shows the last CXR images used as reference, the second row displays the ground-truth CXR images, and the last row showcases the generated CXR images. The comparison between the first and third rows indicates that the generated CXR could well capture anatomical structure, while the comparison between the second and third rows demonstrates that the generated CXRs are in line with the latest imaging manifestations, implying that the disease progression embedded in EHR could be captured and utilized in the generation process. We further retrieve the radiology reports and the discharge summary of the corresponding patients from the database for case studies. \cref{fig:motivating_example} shows one example of the case study (\textit{Sample \#4}), where the patient rapidly turned from normal CXR to severe pulmonary consolidation. The discharge summary shows that the patient experienced transfusion-related acute lung injury and sepsis. Evidently, the generated CXR could more accurately reflect the progressed condition of the patient. Due to space limits, we present more case studies in \cref{app:B2-visual}.

\subsection{Ablation Study}
\begin{table}\small
    \centering
    \caption{Results of the ablation study.}
    \label{tab:ablation_study}
    \begin{tabular}{lcccc}
    \toprule
     &  \multicolumn{2}{c}{Phenotyping} &  \multicolumn{2}{c}{Mortality} \\\midrule
               & AUPRC                                 & AUROC     & AUPRC                                 & AUROC                             \\\midrule
     Last-CXR      & 0.459 {\small$\pm$0.012} & 0.730 {\small$\pm$0.008} & 0.503 {\small$\pm$0.010} & 0.817 {\small$\pm$0.007}      \\\hdashline
     w/o $\mathbf{Z}_{t_0}$       & 0.448 {\small $\pm$0.012} & 0.726 {\small $\pm$0.008} & 0.494 {\small $\pm$0.014} & 0.811 {\small $\pm$0.008}       \\
     w/o $\mathbf{E}_{(t_0, t_1)}$       & 0.461 {\small $\pm$0.002} & 0.723 {\small $\pm$0.006} & 0.474 {\small $\pm$0.016} & 0.799 {\small $\pm$0.015}      \\
     w/o Contrastive    & 0.460 {\small $\pm$0.007} & 0.722 {\small $\pm$0.011} & 0.483 {\small $\pm$0.019} & 0.802 {\small $\pm$0.011}      \\
     w/o $\mathcal{L}_{\text{aux}}$ & 0.461 {\small $\pm$0.003} & 0.718 {\small $\pm$0.012} & 0.495 {\small $\pm$0.026} & 0.811 {\small $\pm$0.009}      \\ \hdashline
     Last-CXR (w/o EHR)  & 0.376 {\small$\pm$0.009} & 0.661 {\small$\pm$0.007} & 0.243 {\small$\pm$0.002} & 0.664 {\small$\pm$0.006}      \\
     \ddlcxr (w/o EHR) & {0.385} {\small$\pm$0.006} & {0.668} {\small$\pm$0.007} & {0.269} {\small$\pm$0.013} & {0.707} {\small$\pm$0.008}\\\hdashline
     \ddlcxr & \textbf{0.470} {\small$\pm$0.003} & \textbf{0.740} {\small$\pm$0.002} & \textbf{0.523} {\small$\pm$0.011} & \textbf{0.822} {\small$\pm$0.009}\\\bottomrule
    \end{tabular}
\end{table}

To better understand the factors contributing to the improved performance, we conducted an ablation study by removing the conditioning components in the LDM stage. The results are summarized in~\cref{tab:ablation_study}.
The variant ``Last-CXR'' has the same architecture as the classifier of \ddlcxr but removes the generated latent CXR $\widehat{\mathbf{Z}}_{t_1}$. The improvement over Last-CXR shows that learning an LDM for generating updated latent CXR is a more effective approach to multimodal fusion, and hence benefits downstream prediction, especially for the mortality prediction task.
The variants ``w/o $\mathbf{Z}_{t_0}$'' and ``w/o $\mathbf{E}_{(t_0, t_1)}$'' remove the last available CXR and EHR data, respectively, from the condition during LDM training. ``w/o Contrastive'' removes the contrastive terms from the LDM objective function. ``w/o $\mathcal{L}_{\text{aux}}$'' removes the auxiliary loss which drives the EHR encoder to capture the CXR-related abnormality findings. The results show that adding each component brings slight improvement while incorporating the reference CXR, the EHR, the contrastive learning, and the auxiliary task achieves the best performance.  {We also remove the EHR data completely in the prediction stage and evaluate the performance using the last available CXR and the generated CXR, respectively. Results are shown as ``Last-CXR (w/o EHR)'' and ``\ddlcxr (w/o EHR)'' in  \cref{tab:ablation_study}. The results suggest that the generation of an updated CXR significantly benefits downstream clinical predictions.} {Additional experiment results on robustness against reduced training data size can be found in \cref{app:abl_reduced_data}.}

\section{Broader Impacts and Limitations}\label{sec:discuss}

\ddlcxr holds promise for societal benefits, such as more precise and timely medical interventions, and offers an alternative for patients with limited access to X-ray imaging. Nonetheless, the potential for generating fake profiles necessitates stringent safeguards, including expert validation of synthesized images, to prevent misuse and protect patient confidentiality, especially when applied to private datasets. Despite its promise, \ddlcxr has some limitations like the need for meticulous hyperparameter tuning and a performance gap across different time intervals, as indicated in Table \ref{tab:mortality_auroc}. Addressing such potential biases is a priority for future research. 
Furthermore, while various metrics have been employed to assess generation quality, expert evaluation by radiologists would provide a more insightful measure of the model's efficacy.

\section{Conclusion}
In this paper, we introduce \ddlcxr, which utilizes a powerful LDM to dynamically generate up-to-date latent chest X-rays to tackle the asynchronicity of multi-modal clinical data for predictions. Our approach involves leveraging various conditions for patient-specific generation: the most recent available chest X-ray to incorporate detailed patient-specific anatomical structure, as well as the EHR data with variable durations for disease progression information. To improve multi-modal fusion in the generation, we develop a contrastive-learning-based LDM to capture and utilize disease courses in EHR. 
Through quantitative and qualitative validations, we demonstrate the superior performance of \ddlcxr in both image generation and enhancing multi-modal fusion via conditional generation for clinical prediction.

\section*{Acknowledgments and Disclosure of Funding}
This work is partially supported by the General Research Fund of Hong Kong Research Grants Council (project no. 15218521), a grant under Theme-based Research Scheme of Hong Kong Research Grants Council (project no. T45-401/22-N), the General Research Fund RGC/HKBU12202621 from the Research Grant Council, the Research Matching Grant Scheme RMGS2021\_8\_06 from the Hong Kong Government, the National Natural
Science Foundation of China (62302413), and the Health and Medical Research Fund (23220312).

\bibliographystyle{unsrtnat}
{\small
\bibliography{references}
}
\newpage
\appendix

\section{Experiment Details}
\label{app:exp}

\subsection{Details of Architectures and Training Procedures of \ddlcxr} \label{app:A1-training}
The training and validation processes are executed on a server equipped with a RTX 4090-24GB GPU card and a 16 vCPU Intel Xeon Processor. The method is implemented using PyTorch 1.9.1 and PyTorch-Lightning 1.4.2. 
DDIM~\cite{ddim} sampling with 200 steps is employed to accelerate the sampling process, and AdamW optimizer is used for all model training. Our implementation is partially based on the repository of the latent diffusion model~\cite{ldm}\footnote{\url{https://github.com/CompVis/latent-diffusion}, open source under MIT license.}. 

\paragraph{Variational autoencoder (VAE)}
The VAE training process, as outlined in~\cite{ldm}, includes a pixel-wise reconstruction loss, a perceptual loss~\cite{Zhang_2018_CVPR}, an adversarial objective, and a lightly-penalized Kullback-Leibler loss toward a standard normal to constrain latent spaces, given by:
\begin{equation}\small
\begin{aligned}
    \mathcal{L}_{\text {VAE}}=\min _{\mathcal{E}, \mathcal{D}, \Phi} \max _\psi\Big(&L_{\text{rec}}(\mathbf{X}^{\text{CXR}}, \mathcal{D}(\mathcal{E}(\mathbf{X}^{\text{CXR}})))-L_{\text{adv}}(\mathcal{D}(\mathcal{E}(\mathbf{X}^{\text{CXR}})))+\log D_\omega(\mathbf{X}^{\text{CXR}}))\\
    &+L_{\text{reg}}(\mathbf{X}^{\text{CXR}}) ; \mathcal{E}, \mathcal{D})+ L_{\text{CE}}(\Phi(\mathcal{E}(\mathbf{X}^{\text{CXR}})), \mathbf{y})\Big),
\end{aligned}
\end{equation}
where $\Phi$ is a classifier that predicts the CheXpert labels associated with the image and we parameterize it with an MLP. All CXRs in the training subset of the LDM stage are gathered for VAE training. A compression rate $r=8$ is adopted, and the training continues for a maximum of 50 epochs. The model with the minimum validation error, as measured using CXRs from the validation subset, is selected. The resulting latent representation has a dimension of $4 \times 28 \times 28 = 3136$. To restrict the normal prior in the latent space and prioritize reconstruction quality, a KL-divergence weighting of $\mathrm{1 \times 10^{-6}}$ is set. We use the base learning rate of $\mathrm{4.5 \times 10^{-6}}$, which is scaled by the number of GPU cards and batch size.

\paragraph{Latent diffusion model (LDM) stage in \ddlcxr}
In the LDM stage of our \ddlcxr model, we employ the UNet architecture~\cite{ronneberger2015u} as the neural backbone, denoted by $\boldsymbol{\epsilon}_\theta$. Meanwhile, we utilize a multivariate time series Transformer~\cite{zerveas2021transformer} for the EHR conditioning encoder $f^{\text{EHR}}_{\text{cond}}$. The Transformer $f^{\text{EHR}}_{\text{cond}}$ is designed with one layer, a model dimension $d$ set to 128, and a maximum EHR data length of 70.
The UNet model $\boldsymbol{\epsilon}_\theta$ features an input channel of 8 and an output channel of 4. The encoding section comprises three blocks, with model channels set at 224, 448, and 672, consisting of a ResBlock module followed by a spatial transformer.  The bottleneck consists of two ResBlock modules with a spatial transformer in between. The decoder mirrors the encoder architecture.
As discussed in Section \ref{sec:ehr_encoder}, we introduce the encoded EHR information through multi-head cross-attention to the spatial transformer module of $\boldsymbol{\epsilon}_{\theta}$. The context dimension is set to 128, and the number of attention heads is 8.
The model is trained for 200 epochs with a batch size of 32, and the model with the smallest composite loss on the validation set is selected for subsequent latent Chest X-ray (CXR) generation. We set the hyperparameters $\alpha$ to 0.2, and $\beta$ to 0.5, empirically.

\paragraph{Latent CXR generation via LDM} In the LDM stage, we aim to generate latent CXR images at time $t_1$, conditioned on $\mathbf{X}^{\text{CXR}}_{t_0}$ and $\mathbf{X}^{\text{EHR}}_{(t_0, t_1)}$. We first encode the CXR images using the pre-trained VAE, given by:
\begin{equation}
    \mathbf{Z}_{t_0} = \mathcal{E}(\mathbf{X}^{\text{CXR}}_{t_0}),\quad \mathbf{Z}_{t_1} = \mathcal{E}(\mathbf{X}^{\text{CXR}}_{t_1}).
\end{equation}

Essentially, the latent CXR generation requires us to estimate the underlying data distribution $q(\mathbf{Z}_{t_1}|\mathbf{Z}_{t_0},\mathbf{X}^{\text{EHR}}_{(t_0,t_1)})$. The LDM approximates this distribution via a model distribution $p_\theta(\mathbf{Z}_{t_1}^{(0)}|\mathbf{Z}_{t_0},\mathbf{X}^{\text{EHR}}_{(t_0,t_1)})$, where $\mathbf{Z}_{t_1}^{(0)} $ represents the prior of a CXR in the latent space encoded by the VAE.

We follow prior work on diffusion models to learn our LDM, which involves two processes~\cite{ddpm, ldm}. In the diffusion process, we gradually add Gaussian noise to $\mathbf{Z}_{t_1}^{(0)}$ in $N$ steps, producing a sequence of noisy representations $\mathbf{Z}_{t_1}^{(1)}, \mathbf{Z}_{t_1}^{(2)},...,\mathbf{Z}_{t_1}^{(N)}$, with the transition probability given by:
\begin{equation}
\begin{aligned}
    q(\mathbf{Z}_{t_1}^{(n)}|\mathbf{Z}_{t_1}^{(n-1)}) &= \mathcal{N} (\mathbf{Z}_{t_1}^{(n)};\sqrt{1-\beta _n} \mathbf{Z}_{t_1}^{n-1} ,\beta_n\mathbf{\mathbf{I}} ),\\
    q(\mathbf{Z}_{t_1}^{(n)}|\mathbf{Z}_{t_1}^{(0)}) &= \mathcal{N} (\mathbf{Z}_{t_1}^{(n)};\sqrt{\bar{\alpha}_n }  \mathbf{Z}_{t_1}^{(0)}, (1-\bar{\alpha}_n)\boldsymbol{\epsilon} ),
\end{aligned}
\end{equation}
where $\boldsymbol{\epsilon} \sim \mathcal{N}(\mathbf{0},\mathbf{I})$ represents the added noise. The noise level is represented by $\bar{\alpha}_n:= {\textstyle \prod_{s=1}^{n}} (1-\beta_s)$, where $\{\beta_n \in (0,1)\}^N_{n=1}$ is a pre-defined variance schedule. At step $N$, $\mathbf{Z}_{t_1}^{(N)}\sim\mathcal{N}(\mathbf{0},\mathbf{I})$ follows an isotropic Gaussian distribution. In the denoising process, we start with a sample from the isotropic Gaussian distribution $\mathbf{Z}_{t_1}^{N}\sim\mathcal{N}(\mathbf{0},\mathbf{I})$ and gradually recreate the latent CXR $\mathbf{Z}_{t_1}^{(0)}$, conditioned on the reference latent CXR, $\mathbf{Z}_{t_0}$, and the EHR data in between, $\mathbf{X}^{\text{EHR}}_{(t_0, t_1)}$. The generation process is given by:
\begin{equation}\small
    p_\theta\left(\mathbf{Z}_{t_1}^{(0:N)}| \mathbf{Z}_{t_0},\mathbf{X}^{\text{EHR}}_{(t_0,t_1)}\right)
=p\left(\mathbf{Z}_{t_1}^{(N)}\right)\prod_{n=1}^{N} p_\theta \left(\mathbf{Z}_{t_1}^{(n-1)} |\mathbf{Z}_{t_1}^{(n)}, \mathbf{Z}_{t_0},\mathbf{X}^{\text{EHR}}_{(t_0,t_1)}\right),
\end{equation}
where
\begin{equation}\small
    p_\theta \left(\mathbf{Z}_{t_1}^{(n-1)} |\mathbf{Z}_{t_1}^{(n)}, \mathbf{Z}_{t_0},\mathbf{X}^{\text{EHR}}_{(t_0,t_1)}\right) 
    =\mathcal{N} \left(\mathbf{Z}_{t_1}^{(n-1)};\boldsymbol{\mu}_\theta\left( 
    \mathbf{Z}_{t_1}^{(n)}, n, \mathbf{Z}_{t_0}, \mathbf{X}_{(t_0,t_1)}^{\text{EHR}}\right),\boldsymbol{\sigma}_n^2\mathbf{I} \right),
\end{equation}

and the mean $\boldsymbol{\mu}_\theta$ and variance $\boldsymbol{\sigma}_n^2$ are parameterized by
\begin{equation}\small
\boldsymbol{\mu}_\theta\left( 
    \mathbf{Z}_{t_1}^{(n)}, n, \mathbf{Z}_{t_0}, \mathbf{X}^{\text{EHR}}_{(t_0,t_1)}\right)
=\frac{1}{\sqrt{\alpha_n}}\left(\mathbf{Z}_{t_1}^{(n)}-\frac{\beta_n}{\sqrt{1-\bar{\alpha}_n}}\boldsymbol{\epsilon}_\theta\left(\mathbf{Z}_{t_1}^{(n)},\mathbf{Z}_{t_0},f^{\text{EHR}}_{\text{cond}}(\mathbf{X}_{(t_0,t_1)}^{\text{EHR}}), n\right)\right),
\label{eqn:ldn_cond_fwd}
\end{equation}
and
\begin{equation}
    \boldsymbol{\sigma}_n^2=\frac{1-\bar{\alpha}_{n-1}}{1-\bar{\alpha}_n}\beta_n,
\end{equation}
where $f^{\text{EHR}}_{\text{cond}}(\cdot)$ is the encoder for the irregular EHR time series. $\boldsymbol{\epsilon}_\theta\left(\mathbf{Z}_{t_1}^{(n)},\mathbf{Z}_{t_0},f^{\text{EHR}}_{\text{cond}}(\mathbf{X}_{(t_0,t_1)}^{\text{EHR}}), n\right)$ denotes the generation noise predicted by the neural backbone.

\paragraph{Prediction stage in \ddlcxr} In the prediction stage, the EHR data is encoded using a one-layer Transformer with a model dimension of 128. We set the dimension of the feedforward layers to 512. The context dimension is also set to 128, and the number of attention heads is 8. We use another Transformer with the same architecture to encode the generated latent CXR $\mathbf{Z}_1$. We use a ResNet-34 model to encode the last available CXR image $\mathbf{X}_0$. The encoded EHR, the encoded latent CXR $\mathbf{Z}_1$, as well as the encoded $\mathbf{X}_0$ are fed into a self-attention layer for final prediction.

\subsection{Details of Data Preprocessing}\label{app:A2-data}

\paragraph{EHR data preprocess}
We follow a similar EHR data extraction and processing pipeline as~\cite{mlhc2022hayatmedfuse} but change the sampling frequency from 2h to 1h and introduce two static variables, age, and gender. We extract 17 clinical time series variables (12 continuous and 5 categorical) with discretization and standardization processes exactly the same as in~\cite{mlhc2022hayatmedfuse}. 
In addition to the 17 clinical time series variables mentioned in the paper ~\cite{mlhc2022hayatmedfuse}, e.g. five categorical (capillary refill rate, Glasgow Coma Scale eye-opening, Glasgow Coma Scale motor response, Glasgow Coma Scale verbal response, and Glasgow Coma Scale total) and 12 continuous (diastolic blood pressure, fraction of inspired oxygen, glucose, heart rate, height, mean blood pressure, oxygen saturation, respiratory rate, systolic blood pressure, temperature, weight, and pH), we introduce two crucial static variables (age and gender) to represent the demographic information of a patient, as patient demographic information is vital for achieving accurate predictions \cite{morid2023time}. To construct our dataset, we sampled time series data at hourly intervals, followed by discretization and standardization processes. We adopt masks to handle missing values in time series to capture the missing pattern, acknowledging that the absence of medical data might be intentional and non-random, driven by caregivers~\cite{morid2023time}.

\paragraph{Data Cohort and Potential Selection Bias} We summarize the number of samples in the LDM stage and the prediction stage in \cref{tab:stats}. The label prevalences of the two prediction tasks are summarized in \cref{tab:Mortality_prevalance,tab:phenotype_label_distribution}. 
We include the disease prevalence in \cref{tab:phenotype_label_distribution} and data cohort statistics in \cref{tab:stats_datacohort}. Here we summarize a few key points:
\begin{itemize}[leftmargin=*, itemsep=0em, parsep=0pt, topsep=0pt, partopsep=0pt]
    \item The statistics of the clinical variables are close to each other, suggesting that patients in our data cohort generally have a similar distribution as that in the entire database.
    \item The overall disease phenotype prevalence is similar.
    \item For a few thorax-related diseases, our data cohort has a slightly higher prevalence, suggesting that potential selection bias exists.
\end{itemize}

\begin{table}[h!]\small
    \centering
    \caption{Data statistics in training, validation, and testing sets for each stage.}
    \begin{tabular}{c c c c}
        \toprule
          \thead{Stage} & \thead{Training} & \thead{Validation} & \thead{Test} \\
         \midrule
         LDM stage & 8545 & 1392 & 2353 \\
         Prediction stage & 5142 & 861 & 1483 \\
         \bottomrule
    \end{tabular}
    \label{tab:stats}
\end{table}

\begin{table}[h!]\small
    \centering
    \caption{Label prevalence of in-hospital mortality prediction task.}
    \begin{tabular}{c c c c}
         \toprule
         & Training & Validation & Test  \\
         \midrule
         Negative & 4396 & 737 & 1264 \\
         Positive & 746 & 124 &219 \\
         Negative/Positive & 5.89 & 5.94 & 5.77\\
         \bottomrule
    \end{tabular}
    \label{tab:Mortality_prevalance}
\end{table}

\begin{table*}[h!]\small
    \centering
    \caption{Number of samples and prevalence of disease phenotypes in training, validation, and testing sets during the prediction stage.  {The prevalence of disease phenotypes among all ICU stays from MIMIC-IV database having \texttt{LoS} $\geq$ \texttt{48h} is given in the last column. }}
     \resizebox{\textwidth}{!}{
    \begin{tabular}{r c c c c c c c}
        \toprule
        \textbf{Disease Label} & \textbf{Training} & \textbf{Validation} & \textbf{Testing} & \textbf{\begin{tabular}[c]{@{}c@{}}Training\\ Prevalence\end{tabular}} & \textbf{\begin{tabular}[c]{@{}c@{}}Validation\\ Prevalence\end{tabular}}  & \textbf{\begin{tabular}[c]{@{}c@{}}Testing\\ Prevalence\end{tabular}} &
        \textbf{\begin{tabular}[c]{@{}c@{}}MIMIC-IV\\ Prevalence\end{tabular}} \\
        \midrule
        Acute and unspecified renal failure & 1932 & 342 & 561 & 0.38 & 0.40 & 0.38& 0.34 \\
        Acute cerebrovascular disease & 467 & 88 & 113 & 0.09 & 0.10 & 0.08 & 0.07\\
        Acute myocardial infarction & 449 & 79 & 131 & 0.09 & 0.09 & 0.09 & 0.09\\
        Cardiac dysrhythmias & 2049 & 361 & 601 & 0.40 & 0.42 & 0.41 & 0.38 \\
        Chronic kidney disease & 1258 & 216 & 399 & 0.24 & 0.25 & 0.27 & 0.23 \\
        Chronic obstructive pulmonary disease  & 860 & 151 & 261 & 0.17 & 0.18 & 0.18 & 0.16\\
        Complications of surgical/medical care & 1218 & 209 & 372 & 0.24 & 0.24 & 0.25 & 0.25\\
        Conduction disorders & 581 & 92 & 176 & 0.11 & 0.11 & 0.12 & 0.12\\
        Congestive heart failure; nonhypertensive & 1674 & 288 & 489 & 0.33 & 0.33 & 0.33 & 0.30 \\
        Coronary atherosclerosis and related & 1605 & 257 & 491 & 0.31 & 0.30 & 0.33 &0.33 \\
        Diabetes mellitus with complications & 646 & 105 & 204 & 0.13 & 0.12 & 0.14 & 0.12 \\
        Diabetes mellitus without complication & 1068 & 182 & 325 & 0.21 & 0.21 & 0.22 & 0.18\\
        Disorders of lipid metabolism & 2047 & 349 & 595 & 0.40 & 0.41 & 0.40 & 0.41 \\
        Essential hypertension & 2255 & 383 & 611 & 0.44 & 0.44 & 0.41 & 0.42\\
        Fluid and electrolyte disorders & 2685 & 455 & 749 & 0.52 & 0.53 & 0.51 & 0.45 \\
        Gastrointestinal hemorrhage & 354 & 68 & 129 & 0.07 & 0.08 & 0.09 & 0.07\\
        Hypertension with complications & 1129 & 202 & 361 & 0.22 & 0.23 & 0.24 &0.24 \\
        Other liver diseases & 899 & 147 & 279 & 0.17 & 0.17 & 0.19 & 0.15\\
        Other lower respiratory disease & 718 & 109 & 212 & 0.14 & 0.13 & 0.14  & 0.12\\
        Other upper respiratory disease & 376 & 67 & 96 & 0.07 & 0.08 & 0.06 &0.07\\
        Pleurisy; pneumothorax; pulmonary collapse & 553 & 99 & 141 & 0.11 & 0.11 & 0.10 & 0.09 \\
        Pneumonia & 1149 & 187 & 333 & 0.22 & 0.22 & 0.22 &0.18 \\
        Respiratory failure; insufficiency; arrest (adult) & 1741 & 307 & 506 & 0.34 & 0.36 & 0.34  &0.25\\
        Septicemia (except in labor) & 1386 & 224 & 425 & 0.27 & 0.26 & 0.29 &0.21 \\
        Shock & 1157 & 192 & 357 & 0.23 & 0.22 & 0.24 &0.18 \\
        \bottomrule
    \end{tabular}
    }
    \label{tab:phenotype_label_distribution}
\end{table*}

\begin{table}[h!]\small
    \centering 
    \caption{Statistics of data cohort. We report mean±std for continuous variables and the mode for categorical variables.}
    \scalebox{0.85}{
    \begin{tabular}{r c c}
        \toprule
        \textbf{Variables} & 
        \textbf{\begin{tabular}[c]{@{}c@{}}DDL-CXR\\ Cohort\end{tabular}}  & 
        \textbf{\begin{tabular}[c]{@{}c@{}}MIMIC-IV\\ Database\end{tabular}} \\
         \midrule
            Sample size & 5142 & 26330\\
            Age & 63.9±16.6 & 64.3±16.4\\
            Gender (Men, \%) & 54.98 & 55.99\\
            Diastolic blood pressure & 62.2±10.1 & 62.8±10.5\\
            Fraction inspired oxygen & 0.4±0.2 & 0.4±0.2\\
            Glucose & 141.1±41.7 & 139±39.6\\
            Heart Rate & 87.3±15.4 & 85.6±15\\
            Height & 170±1.1 & 170±1.5\\
            Mean blood pressure & 77.1±10 & 78.3±10.4\\
            Oxygen saturation & 96.7±2 & 96.5±2\\
            Respiratory rate & 19.8±3.7 & 19.5±3.7\\
            Systolic blood pressure & 118.3±15.2 & 118.6±15.6\\
            Temperature & 37±0.4 & 36.9±0.4\\
            Weight & 80.8±20 & 81.5±19.5\\
            pH & 7.2±0.3 & 7.2±0.3\\
            Capillary refill rate & Normal & Normal\\
            GCS - eye opening & Spontaneously & Spontaneously\\
            GCS - motor response & Obeys Commands & Obeys Commands\\
            GCS – verbal response & Oriented & Oriented\\
            GCS - total & 15 & 15\\
            In-hospital mortality (\%) & 15 & 12\\
         \bottomrule
    \end{tabular}
    }
    \label{tab:stats_datacohort}
\end{table}

\clearpage
\section{Additional Results}

\subsection{AUPRC of Mortality Prediction}\label{app:B1-AUPRC}
We summarize the AUPRC score for the mortality prediction task in \cref{tab:mortality_auprc}.
\begin{table*}[h!]\small
    \centering
    \caption{The AUPRC score for mortality prediction for overall and different time gaps. $\delta$ represents the time gap between the generation (or prediction) time and the occurrence time of the last available CXR. Numbers in bold indicate the best performance in each column. \ddlcxr outperforms all baselines in overall and $\delta <$ 24h settings.}\label{tab:mortality_auprc}
    \begin{tabular}{cccccc}
    \toprule
 & Overall & $\delta<12$ & $12 \leq\delta <24$ & $24 \leq\delta <36$ & $\delta \geq 36$\\
\textit{prevalence} & 14.7\% & 16.6\% & 19\% & 15.9\% & 9.26\%\\\midrule
Uni-EHR~\cite{vaswani2017attention} & 0.498 {\small$\pm$0.007} & 0.579 {\small$\pm$0.009} & 0.541 {\small$\pm$0.016} & 0.416 {\small$\pm$0.018} & 0.411 {\small$\pm$0.005}\\
MMTM~\cite{joze2020mmtm} & 0.422 {\small$\pm$0.014} & 0.430 {\small$\pm$0.010} & 0.458 {\small$\pm$0.020} & 0.476 {\small$\pm$0.021} & 0.374 {\small$\pm$0.012}\\
DAFT~\cite{daft-polsterl2021combining} & 0.448 {\small$\pm$0.004} & 0.460 {\small$\pm$0.011} & 0.478 {\small$\pm$0.010} & \textbf{0.508} {\small$\pm$0.015} & 0.395 {\small$\pm$0.007}\\
MedFuse~\cite{mlhc2022hayatmedfuse} & 0.443 {\small$\pm$0.009} & 0.498 {\small$\pm$0.013} & 0.484 {\small$\pm$0.023} & 0.432 {\small$\pm$0.015} & 0.355 {\small$\pm$0.008}\\
DrFuse~\cite{yao2024drfuse} & 0.460 {\small$\pm$0.004} & 0.506 {\small$\pm$0.033} & 0.356 {\small$\pm$0.034} & 0.382 {\small$\pm$0.072} & 0.399 {\small$\pm$0.019}\\
GAN-based~\cite{xia2021learning} & 0.505 {\small$\pm$0.018} & 0.567 {\small$\pm$0.015} & 0.551 {\small$\pm$0.018} & 0.414 {\small$\pm$0.041} & \textbf{0.442} {\small$\pm$0.034}\\
\ddlcxr (ours) & \textbf{0.523} {\small$\pm$0.011} & \textbf{0.610} {\small$\pm$0.015} & \textbf{0.566} {\small$\pm$0.013} & 0.429 {\small$\pm$0.009} & 0.440 {\small$\pm$0.008}\\\bottomrule
    \end{tabular}
\end{table*}

\subsection{AUROC of Phenotype Prediction by Disease Label}\label{app:B13-disease}
We summarize the AUROC score for the phenotype prediction task by disease label in \cref{tab:disease_auroc}.
\begin{table}[h!]
\setlength{\tabcolsep}{3pt}
    \centering
    \caption{The AUROC score by disease labels. Results show that \ddlcxr effectively tackles the asynchronicity issue in clinical multi-modal fusion, achieving the highest average rank across all disease labels.}\label{tab:disease_auroc}
    \resizebox{\linewidth}{!}{
    \begin{tabular}{lccccccc}\toprule
 & Uni-EHR~\cite{vaswani2017attention} & MMTM~\cite{joze2020mmtm} & DAFT~\cite{daft-polsterl2021combining} & MedFuse~\cite{mlhc2022hayatmedfuse} & DrFuse~\cite{yao2024drfuse} & GAN-based~\cite{xia2021learning} & \ddlcxr (ours)\\\midrule
Acute renal failure & \textit{0.718}{\small $\pm$0.003} & 0.706{\small $\pm$0.004} & 0.712{\small $\pm$0.003} & 0.706{\small $\pm$0.004} & 0.702{\small $\pm$0.003} & 0.710{\small $\pm$0.005} & \textbf{0.723}{\small $\pm$0.004}\\
Acute cerebrovascular disease & 0.880{\small $\pm$0.004} & \textit{0.893}{\small $\pm$0.003} & 0.888{\small $\pm$0.009} & \textbf{0.895}{\small $\pm$0.005} & 0.873{\small $\pm$0.009} & 0.875{\small $\pm$0.011} & 0.870{\small $\pm$0.008}\\
Acute myocardial infarction & 0.712{\small $\pm$0.006} & 0.717{\small $\pm$0.005} & 0.722{\small $\pm$0.009} & 0.724{\small $\pm$0.009} & \textit{0.730}{\small $\pm$0.013} & 0.701{\small $\pm$0.004} & \textbf{0.735}{\small $\pm$0.009}\\
Cardiac dysrhythmias & 0.677{\small $\pm$0.005} & 0.648{\small $\pm$0.004} & 0.656{\small $\pm$0.004} & 0.667{\small $\pm$0.010} & \textit{0.682}{\small $\pm$0.006} & 0.673{\small $\pm$0.005} & \textbf{0.703}{\small $\pm$0.007}\\
Chronic kidney disease & 0.745{\small $\pm$0.008} & 0.735{\small $\pm$0.008} & 0.732{\small $\pm$0.004} & 0.720{\small $\pm$0.004} & 0.710{\small $\pm$0.011} & \textit{0.748}{\small $\pm$0.005} & \textbf{0.768}{\small $\pm$0.006}\\
COPD and bronchiectasis & 0.706{\small $\pm$0.006} & 0.689{\small $\pm$0.014} & 0.707{\small $\pm$0.012} & 0.686{\small $\pm$0.006} & 0.733{\small $\pm$0.011} & \textit{0.738}{\small $\pm$0.003} & \textbf{0.740}{\small $\pm$0.003}\\
Surgical complications & 0.636{\small $\pm$0.007} & \textit{0.656}{\small $\pm$0.008} & 0.648{\small $\pm$0.006} & \textbf{0.662}{\small $\pm$0.008} & 0.649{\small $\pm$0.004} & 0.610{\small $\pm$0.013} & 0.649{\small $\pm$0.003}\\
Conduction disorders & 0.741{\small $\pm$0.006} & 0.717{\small $\pm$0.011} & 0.735{\small $\pm$0.012} & 0.724{\small $\pm$0.008} & \textit{0.847}{\small $\pm$0.011} & 0.842{\small $\pm$0.006} & \textbf{0.860}{\small $\pm$0.005}\\
CHF; nonhypertensive & 0.771{\small $\pm$0.004} & 0.777{\small $\pm$0.010} & 0.796{\small $\pm$0.007} & 0.779{\small $\pm$0.008} & 0.797{\small $\pm$0.004} & \textit{0.798}{\small $\pm$0.006} & \textbf{0.820}{\small $\pm$0.005}\\
Coronary atherosclerosis \& others & 0.740{\small $\pm$0.005} & 0.728{\small $\pm$0.003} & 0.733{\small $\pm$0.007} & 0.721{\small $\pm$0.008} & 0.742{\small $\pm$0.006} & \textit{0.753}{\small $\pm$0.007} & \textbf{0.764}{\small $\pm$0.006}\\
DM with complications & \textbf{0.858}{\small $\pm$0.006} & 0.853{\small $\pm$0.006} & 0.851{\small $\pm$0.002} & 0.845{\small $\pm$0.005} & 0.849{\small $\pm$0.006} & 0.851{\small $\pm$0.005} & \textit{0.855}{\small $\pm$0.004}\\
DM without complication & \textit{0.716}{\small $\pm$0.005} & 0.707{\small $\pm$0.007} & 0.694{\small $\pm$0.015} & 0.704{\small $\pm$0.003} & 0.712{\small $\pm$0.008} & 0.707{\small $\pm$0.008} & \textbf{0.717}{\small $\pm$0.009}\\
Disorders of lipid metabolism & 0.704{\small $\pm$0.007} & 0.677{\small $\pm$0.005} & 0.676{\small $\pm$0.006} & 0.677{\small $\pm$0.003} & 0.696{\small $\pm$0.008} & \textit{0.708}{\small $\pm$0.006} & \textbf{0.714}{\small $\pm$0.007}\\
Essential hypertension & \textit{0.662}{\small $\pm$0.007} & 0.609{\small $\pm$0.002} & 0.625{\small $\pm$0.005} & 0.632{\small $\pm$0.006} & 0.612{\small $\pm$0.003} & 0.659{\small $\pm$0.008} & \textbf{0.667}{\small $\pm$0.008}\\
Fluid and electrolyte disorders & 0.671{\small $\pm$0.002} & 0.674{\small $\pm$0.003} & 0.671{\small $\pm$0.003} & 0.665{\small $\pm$0.006} & 0.671{\small $\pm$0.004} & \textit{0.674}{\small $\pm$0.007} & \textbf{0.677}{\small $\pm$0.007}\\
Gastrointestinal hemorrhage & 0.664{\small $\pm$0.005} & 0.635{\small $\pm$0.017} & 0.653{\small $\pm$0.018} & 0.664{\small $\pm$0.005} & \textit{0.677}{\small $\pm$0.013} & 0.662{\small $\pm$0.008} & \textbf{0.678}{\small $\pm$0.009}\\
Secondary hypertension & 0.725{\small $\pm$0.007} & 0.726{\small $\pm$0.007} & 0.722{\small $\pm$0.005} & 0.716{\small $\pm$0.004} & 0.714{\small $\pm$0.010} & \textit{0.734}{\small $\pm$0.005} & \textbf{0.752}{\small $\pm$0.004}\\
Other liver diseases & 0.681{\small $\pm$0.007} & 0.680{\small $\pm$0.005} & 0.693{\small $\pm$0.006} & 0.690{\small $\pm$0.004} & 0.697{\small $\pm$0.009} & \textit{0.700}{\small $\pm$0.010} & \textbf{0.708}{\small $\pm$0.007}\\
Other lower respiratory disease & 0.596{\small $\pm$0.004} & 0.606{\small $\pm$0.007} & 0.608{\small $\pm$0.010} & 0.615{\small $\pm$0.006} & \textbf{0.634}{\small $\pm$0.015} & \textit{0.619}{\small $\pm$0.019} & 0.614{\small $\pm$0.011}\\
Other upper respiratory disease & 0.691{\small $\pm$0.009} & 0.691{\small $\pm$0.009} & 0.711{\small $\pm$0.006} & \textit{0.726}{\small $\pm$0.010} & \textbf{0.741}{\small $\pm$0.013} & 0.699{\small $\pm$0.019} & 0.700{\small $\pm$0.017}\\
Pleurisy; pneumothorax & 0.634{\small $\pm$0.005} & 0.647{\small $\pm$0.013} & \textit{0.674}{\small $\pm$0.014} & 0.673{\small $\pm$0.008} & \textbf{0.687}{\small $\pm$0.014} & 0.670{\small $\pm$0.027} & 0.666{\small $\pm$0.010}\\
Pneumonia & 0.704{\small $\pm$0.006} & \textbf{0.725}{\small $\pm$0.004} & \textit{0.723}{\small $\pm$0.007} & 0.714{\small $\pm$0.005} & 0.715{\small $\pm$0.008} & 0.707{\small $\pm$0.004} & 0.719{\small $\pm$0.008}\\
Respiratory failure & 0.800{\small $\pm$0.005} & \textbf{0.816}{\small $\pm$0.005} & \textit{0.815}{\small $\pm$0.003} & 0.809{\small $\pm$0.005} & 0.810{\small $\pm$0.004} & 0.805{\small $\pm$0.004} & 0.811{\small $\pm$0.008}\\
Septicemia (except in labor) & \textit{0.764}{\small $\pm$0.005} & 0.757{\small $\pm$0.005} & 0.760{\small $\pm$0.007} & 0.741{\small $\pm$0.005} & 0.749{\small $\pm$0.003} & 0.763{\small $\pm$0.004} & \textbf{0.774}{\small $\pm$0.005}\\
Shock & 0.803{\small $\pm$0.004} & 0.803{\small $\pm$0.005} & 0.805{\small $\pm$0.006} & \textbf{0.807}{\small $\pm$0.003} & 0.800{\small $\pm$0.008} & 0.801{\small $\pm$0.007} & \textit{0.806}{\small $\pm$0.007}\\\midrule
\multicolumn{1}{r}{Average Rank} & 4.56 & 4.76 & 4.32 & 4.64 & 3.88 & 3.84 & 2\\\bottomrule
\end{tabular}}
\end{table}

\clearpage
\subsection{AUPRC of Phenotype Prediction by Disease Label}\label{app:B14-disease-auprc-sup}
We summarize the AUPRC score for the phenotype prediction task by disease label in \cref{tab:disease_auprc_sup}.
\begin{table}[h!]
\setlength{\tabcolsep}{3pt}
    \centering
    \caption{The AUPRC score by disease labels with detailed standard deviation.}\label{tab:disease_auprc_sup}
    \resizebox{\linewidth}{!}{
    \begin{tabular}{lccccccc}\toprule
 & Uni-EHR~\cite{vaswani2017attention} & MMTM~\cite{joze2020mmtm} & DAFT~\cite{daft-polsterl2021combining} & MedFuse~\cite{mlhc2022hayatmedfuse} & DrFuse~\cite{yao2024drfuse} & GAN-based~\cite{xia2021learning} & \ddlcxr (ours)\\\midrule
Acute renal failure & \textit{0.573}{\small $\pm$0.004} & 0.568{\small $\pm$0.007} & 0.572{\small $\pm$0.006} & 0.565{\small $\pm$0.005} & 0.564{\small $\pm$0.005} & 0.563{\small $\pm$0.005} & \textbf{0.588}{\small $\pm$0.008}\\
Acute cerebrovascular disease & 0.425{\small $\pm$0.010} & 0.418{\small $\pm$0.010} & 0.419{\small $\pm$0.017} & \textit{0.434}{\small $\pm$0.017} & 0.399{\small $\pm$0.019} & \textbf{0.446}{\small $\pm$0.010} & 0.416{\small $\pm$0.018}\\
Acute myocardial infarction & 0.185{\small $\pm$0.008} & 0.192{\small $\pm$0.004} & 0.187{\small $\pm$0.009} & \textbf{0.219}{\small $\pm$0.015} & \textit{0.209}{\small $\pm$0.013} & 0.171{\small $\pm$0.008} & 0.206{\small $\pm$0.020}\\
Cardiac dysrhythmias & 0.579{\small $\pm$0.006} & 0.532{\small $\pm$0.015} & 0.548{\small $\pm$0.005} & 0.560{\small $\pm$0.012} & \textit{0.584}{\small $\pm$0.009} & 0.561{\small $\pm$0.008} & \textbf{0.605}{\small $\pm$0.009}\\
Chronic kidney disease & 0.515{\small $\pm$0.016} & 0.505{\small $\pm$0.011} & \textit{0.515}{\small $\pm$0.009} & 0.497{\small $\pm$0.009} & 0.477{\small $\pm$0.016} & 0.501{\small $\pm$0.006} & \textbf{0.538}{\small $\pm$0.010}\\
COPD and bronchiectasis & 0.319{\small $\pm$0.012} & 0.327{\small $\pm$0.021} & 0.342{\small $\pm$0.011} & 0.344{\small $\pm$0.004} & \textbf{0.405}{\small $\pm$0.010} & 0.372{\small $\pm$0.012} & \textit{0.382}{\small $\pm$0.004}\\
Surgical complications & 0.370{\small $\pm$0.009} & 0.379{\small $\pm$0.006} & \textit{0.385}{\small $\pm$0.007} & 0.381{\small $\pm$0.006} & 0.377{\small $\pm$0.003} & 0.344{\small $\pm$0.006} & \textbf{0.388}{\small $\pm$0.004}\\
Conduction disorders & 0.276{\small $\pm$0.006} & 0.287{\small $\pm$0.007} & 0.298{\small $\pm$0.013} & 0.286{\small $\pm$0.016} & \textit{0.632}{\small $\pm$0.010} & 0.609{\small $\pm$0.004} & \textbf{0.633}{\small $\pm$0.003}\\
CHF; nonhypertensive & 0.593{\small $\pm$0.007} & 0.619{\small $\pm$0.019} & 0.647{\small $\pm$0.017} & 0.631{\small $\pm$0.011} & \textit{0.661}{\small $\pm$0.015} & 0.652{\small $\pm$0.012} & \textbf{0.682}{\small $\pm$0.009}\\
CAD & 0.560{\small $\pm$0.007} & 0.540{\small $\pm$0.006} & 0.556{\small $\pm$0.008} & 0.544{\small $\pm$0.009} & 0.581{\small $\pm$0.012} & \textit{0.590}{\small $\pm$0.014} & \textbf{0.611}{\small $\pm$0.010}\\
DM with complications & \textit{0.562}{\small $\pm$0.013} & \textbf{0.569}{\small $\pm$0.016} & 0.552{\small $\pm$0.006} & 0.561{\small $\pm$0.012} & 0.550{\small $\pm$0.015} & 0.552{\small $\pm$0.019} & 0.524{\small $\pm$0.007}\\
DM without complication & \textbf{0.370}{\small $\pm$0.005} & 0.367{\small $\pm$0.013} & 0.343{\small $\pm$0.014} & 0.356{\small $\pm$0.006} & \textit{0.369}{\small $\pm$0.010} & 0.352{\small $\pm$0.012} & 0.368{\small $\pm$0.008}\\
Disorders of lipid metabolism & \textit{0.594}{\small $\pm$0.007} & 0.576{\small $\pm$0.002} & 0.570{\small $\pm$0.009} & 0.566{\small $\pm$0.003} & 0.584{\small $\pm$0.010} & 0.587{\small $\pm$0.008} & \textbf{0.601}{\small $\pm$0.010}\\
Essential hypertension & 0.551{\small $\pm$0.006} & 0.519{\small $\pm$0.003} & 0.525{\small $\pm$0.005} & 0.518{\small $\pm$0.009} & 0.502{\small $\pm$0.005} & \textit{0.554}{\small $\pm$0.011} & \textbf{0.561}{\small $\pm$0.011}\\
Fluid and electrolyte disorders & 0.655{\small $\pm$0.003} & \textit{0.664}{\small $\pm$0.005} & 0.662{\small $\pm$0.004} & 0.656{\small $\pm$0.008} & 0.658{\small $\pm$0.006} & 0.662{\small $\pm$0.012} & \textbf{0.672}{\small $\pm$0.008}\\
Gastrointestinal hemorrhage & 0.180{\small $\pm$0.013} & 0.142{\small $\pm$0.014} & 0.162{\small $\pm$0.019} & \textbf{0.192}{\small $\pm$0.014} & \textit{0.191}{\small $\pm$0.014} & 0.151{\small $\pm$0.008} & 0.180{\small $\pm$0.009}\\
Secondary hypertension & \textit{0.463}{\small $\pm$0.012} & 0.455{\small $\pm$0.007} & 0.452{\small $\pm$0.007} & 0.453{\small $\pm$0.013} & 0.437{\small $\pm$0.012} & 0.451{\small $\pm$0.011} & \textbf{0.484}{\small $\pm$0.009}\\
Other liver diseases & 0.316{\small $\pm$0.013} & 0.316{\small $\pm$0.010} & 0.341{\small $\pm$0.007} & 0.344{\small $\pm$0.008} & \textit{0.372}{\small $\pm$0.014} & 0.362{\small $\pm$0.018} & \textbf{0.378}{\small $\pm$0.009}\\
Other lower respiratory disease & 0.219{\small $\pm$0.003} & 0.209{\small $\pm$0.012} & 0.206{\small $\pm$0.008} & 0.223{\small $\pm$0.005} & \textbf{0.255}{\small $\pm$0.011} & 0.236{\small $\pm$0.008} & \textit{0.242}{\small $\pm$0.007}\\
Other upper respiratory disease & 0.166{\small $\pm$0.015} & 0.137{\small $\pm$0.009} & 0.166{\small $\pm$0.019} & 0.202{\small $\pm$0.014} & \textbf{0.274}{\small $\pm$0.018} & 0.196{\small $\pm$0.019} & \textit{0.234}{\small $\pm$0.059}\\
Pleurisy; pneumothorax & 0.143{\small $\pm$0.005} & 0.145{\small $\pm$0.008} & 0.159{\small $\pm$0.009} & 0.159{\small $\pm$0.013} & \textbf{0.172}{\small $\pm$0.007} & \textit{0.171}{\small $\pm$0.022} & 0.166{\small $\pm$0.009}\\
Pneumonia & 0.412{\small $\pm$0.012} & \textbf{0.437}{\small $\pm$0.005} & \textit{0.429}{\small $\pm$0.008} & 0.419{\small $\pm$0.008} & 0.406{\small $\pm$0.016} & 0.415{\small $\pm$0.010} & 0.428{\small $\pm$0.013}\\
Respiratory failure & 0.655{\small $\pm$0.009} & \textit{0.686}{\small $\pm$0.011} & 0.674{\small $\pm$0.004} & 0.671{\small $\pm$0.009} & \textbf{0.692}{\small $\pm$0.005} & 0.663{\small $\pm$0.007} & 0.669{\small $\pm$0.012}\\
Septicemia (except in labor) & \textit{0.585}{\small $\pm$0.008} & 0.573{\small $\pm$0.012} & 0.580{\small $\pm$0.014} & 0.565{\small $\pm$0.009} & 0.562{\small $\pm$0.013} & 0.573{\small $\pm$0.009} & \textbf{0.603}{\small $\pm$0.010}\\
Shock & \textit{0.590}{\small $\pm$0.002} & 0.584{\small $\pm$0.005} & 0.582{\small $\pm$0.009} & \textbf{0.592}{\small $\pm$0.008} & 0.572{\small $\pm$0.021} & 0.587{\small $\pm$0.011} & 0.586{\small $\pm$0.013}\\\midrule
\multicolumn{1}{r}{Average Rank} & 4.4 & 4.64 & 4.4 & 4.24 & 3.88 & 4.16 & 2.28\\\bottomrule
\end{tabular}}
\end{table}

\subsection{Ablation Study for a Reduced Percentage of Data}\label{app:abl_reduced_data}
To investigate the sensitivity to the amount of training data of the proposed model and the main baselines, we conduct experiments with varying training sizes. The results are summarized in \cref{tab:ablation_study_red_data_pheno} and \cref{tab:ablation_study_red_data_mort}. Results show that DDL-CXR consistently outperforms baseline methods by a large margin, demonstrating its robustness against training data size.
\begin{table}[h!]\small
    \centering
    \caption{Performance of the phenotype classification task with different training sizes. Results are reported in mean±std.}
    \label{tab:ablation_study_red_data_pheno}
    \resizebox{\linewidth}{!}{
    \begin{tabular}{ccccccc}
    \toprule
     &  \multicolumn{3}{c}{AUPRC} &  \multicolumn{3}{c}{AUROC} \\\midrule
               & 100\% & 75\%     & 50\%         & 100\% & 75\%     & 50\%    \\\midrule
Uni-EHR~\cite{vaswani2017attention} & 0.434 {\small$\pm$0.009} & 0.428 {\small$\pm$0.011} & 0.419 {\small$\pm$0.010} & 0.720 {\small$\pm$0.006} & 0.711 {\small$\pm$0.008} & 0.706 {\small$\pm$0.006} \\
MMTM~\cite{joze2020mmtm} & 0.430 {\small$\pm$0.005} & 0.421 {\small$\pm$0.004} & 0.406 {\small$\pm$0.003} & 0.715 {\small$\pm$0.003} & 0.707 {\small$\pm$0.002} & 0.694 {\small$\pm$0.002} \\
DAFT~\cite{daft-polsterl2021combining} & 0.435 {\small$\pm$0.002} & 0.422 {\small$\pm$0.004} & 0.407 {\small$\pm$0.003} & 0.720 {\small$\pm$0.003} & 0.709 {\small$\pm$0.002} & 0.699 {\small$\pm$0.003} \\
MedFuse~\cite{mlhc2022hayatmedfuse} & 0.437 {\small$\pm$0.001} & 0.420 {\small$\pm$0.004} & 0.412 {\small$\pm$0.002} & 0.718 {\small$\pm$0.002} & 0.707 {\small$\pm$0.003} & 0.700 {\small$\pm$0.001} \\
DrFuse~\cite{yao2024drfuse} & 0.459 {\small$\pm$0.003} & 0.442 {\small$\pm$0.005} & 0.431 {\small$\pm$0.004} & 0.729 {\small$\pm$0.004} & 0.717 {\small$\pm$0.005} & 0.709 {\small$\pm$0.004} \\
\ddlcxr & \textbf{0.470} {\small$\pm$0.003} & \textbf{0.457} {\small$\pm$0.003} & \textbf{0.433} {\small$\pm$0.011} & \textbf{0.740} {\small$\pm$0.002} & \textbf{0.734} {\small$\pm$0.002} & \textbf{0.715} {\small$\pm$0.005} \\\bottomrule
    \end{tabular}}
\end{table}

\begin{table}[h!]\small
    \centering
    \caption{Performance of the mortality prediction task with different training sizes. Results are reported in mean±std.}
    \label{tab:ablation_study_red_data_mort}
    \resizebox{\linewidth}{!}{
    \begin{tabular}{ccccccc}
    \toprule
     &  \multicolumn{3}{c}{AUPRC} &  \multicolumn{3}{c}{AUROC} \\\midrule
               & 100\% & 75\%     & 50\%         & 100\% & 75\%     & 50\%    \\\midrule
Uni-EHR~\cite{vaswani2017attention} & 0.498 {\small$\pm$0.007} & 0.437 {\small$\pm$0.011} & 0.429 {\small$\pm$0.012} & 0.815 {\small$\pm$0.007} & 0.791 {\small$\pm$0.005} & 0.782 {\small$\pm$0.013} \\
MMTM~\cite{joze2020mmtm} & 0.422 {\small$\pm$0.014} & 0.405 {\small$\pm$0.010} & 0.399 {\small$\pm$0.012} & 0.785 {\small$\pm$0.004} & 0.782 {\small$\pm$0.002} & 0.775 {\small$\pm$0.005}  \\
DAFT~\cite{daft-polsterl2021combining} & 0.448 {\small$\pm$0.004} & 0.428 {\small$\pm$0.006} & 0.413 {\small$\pm$0.005} & 0.800 {\small$\pm$0.003} & 0.790 {\small$\pm$0.004} & 0.778 {\small$\pm$0.007}  \\
MedFuse~\cite{mlhc2022hayatmedfuse} & 0.443 {\small$\pm$0.009} & 0.420 {\small$\pm$0.015} & 0.411 {\small$\pm$0.009} & 0.793 {\small$\pm$0.003} & 0.784 {\small$\pm$0.004} & 0.775 {\small$\pm$0.005}  \\
DrFuse~\cite{yao2024drfuse} & 0.460 {\small$\pm$0.004} & 0.430 {\small$\pm$0.013} & 0.415 {\small$\pm$0.030} & 0.773 {\small$\pm$0.008} & 0.755 {\small$\pm$0.008} & 0.766 {\small$\pm$0.030} \\
\ddlcxr & \textbf{0.523} {\small$\pm$0.011} & \textbf{0.474} {\small$\pm$0.009} & \textbf{0.466} {\small$\pm$0.012} & \textbf{0.822} {\small$\pm$0.009} & \textbf{0.801} {\small$\pm$0.008} & \textbf{0.790} {\small$\pm$0.008}  \\\bottomrule
    \end{tabular}}
\end{table}

\clearpage
\subsection{Additional Case Studies}\label{app:B2-visual}
We demonstrate additional case studies in \cref{fig:sample1} -- \cref{fig:sample7}. We retrieve findings from the radiology reports. The case studies show that \ddlcxr could generate CXR images that align with the disease progression of the individual patient.

\newcolumntype{C}[1]{>{\centering\arraybackslash}m{#1}}

\begin{figure}[h!]\small 
    \centering
    \scalebox{0.85}{
    \begin{tabular}{C{15em} C{15em} C{15em}} 
        \includegraphics[width=0.9\linewidth, valign=c]{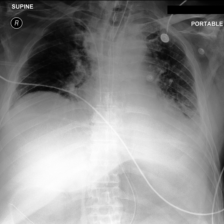}
        & \includegraphics[width=0.9\linewidth, valign=c]{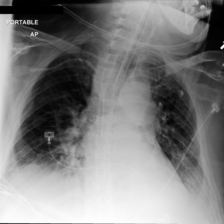}
        & \includegraphics[width=0.9\linewidth, valign=c]{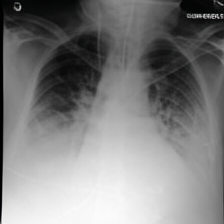}\\
    \end{tabular}
    }
    \scalebox{0.85}{
    \begin{tabular}{p{15em} p{15em} p{15em}}
        {\small(a)\textit{Initial radiology findings}: the chest demonstrates low lung volumes,
 which results in bronchovascular crowding. Bibasilar opacities may reflect
 atelectasis There is a probable small left pleural effusion.
        }
         & {\small(b) \textit{Radiology findings after 8 hours}: There is interval progression of interstitial pulmonary edema and
 potential interval increase in bibasal consolidations.}
         & {\small (c) \small CXR image generated by \ddlcxr given the initial CXR image shown in (a) and the EHR data within the 8 hours.}
    \end{tabular}
    }
     \caption{Case Study of Sample \#1}\label{fig:sample1}
\end{figure}

\begin{figure}[h!]\small
    \centering
    \scalebox{0.85}{
    \begin{tabular}{C{15em} C{15em} C{15em}} 
        \includegraphics[width=0.9\linewidth, valign=c]{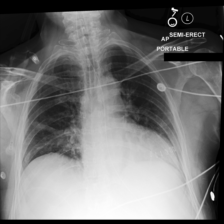}
        & \includegraphics[width=0.9\linewidth, valign=c]{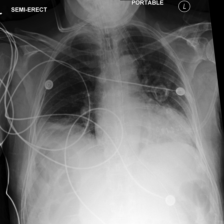}
        & \includegraphics[width=0.9\linewidth, valign=c]{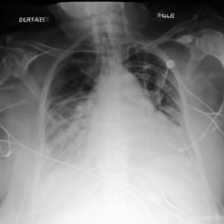}\\
    \end{tabular}
    }
    \scalebox{0.85}{
    \begin{tabular}{p{15em} p{15em} p{15em}}
        {\small(a)  Heart remains enlarged with left ventricular prominence.  Interval
 appearance of linear opacity in the right mid lung which may reflect fluid
 loculated within the minor fissure or possibly subsegmental atelectasis.
  }
         & {\small(b) \textit{Radiology findings after 18 hours}:  
 There are fluctuating patchy opacities at the right lung
 base suggestive of atelectasis.  Low volumes with crowding of the
 pulmonary vasculature.}
         & {\small (c) \small CXR image generated by \ddlcxr given the initial CXR image shown in (a) and the EHR data within the 18 hours.}
    \end{tabular}
    }
     \caption{Case Study of Sample \#2}\label{fig:sample2}
\end{figure}

\begin{figure}[h!]\small
    \centering
    \scalebox{0.85}{
    \begin{tabular}{C{15em} C{15em} C{15em}} 
        \includegraphics[width=0.9\linewidth, valign=c]{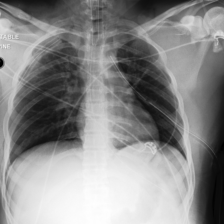}
        & \includegraphics[width=0.9\linewidth, valign=c]{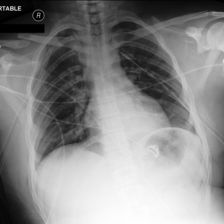}
        & \includegraphics[width=0.9\linewidth, valign=c]{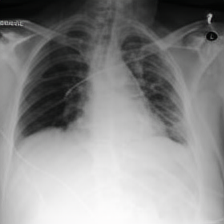}\\
    \end{tabular}
    }
    \scalebox{0.85}{
    \begin{tabular}{p{15em} p{15em} p{15em}}
        {\small(a)  The dense atelectatic streaks in the
 left mid zone has decreased. The bilateral chest tubes remain in place and
 there is no evidence of pneumothorax. }
         & {\small(b) \textit{Radiology findings after 13 hours}: No left pneumothorax
 is appreciated.
 The hemidiaphragms are now sharp be seen with only mild atelectatic changes at
 the bases.}
         & {\small (c) \small CXR image generated by \ddlcxr given the initial CXR image shown in (a) and the EHR data within the 13 hours.}
    \end{tabular}
    }
     \caption{Case Study of Sample \#3}\label{fig:sample3}
\end{figure}

\begin{figure}[h!]\small
    \centering
    \scalebox{0.85}{
    \begin{tabular}{C{15em} C{15em} C{15em}} 
        \includegraphics[width=0.9\linewidth, valign=c]{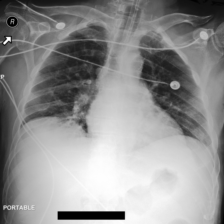}
        & \includegraphics[width=0.9\linewidth, valign=c]{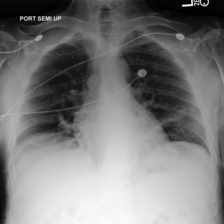}
        & \includegraphics[width=0.9\linewidth, valign=c]{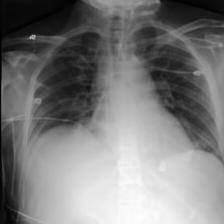}\\
    \end{tabular}
    }
    \scalebox{0.85}{
    \begin{tabular}{p{15em} p{15em} p{15em}}
        {\small(a) \textit{Initial radiology findings}:The lung
 volumes are low.  Mild cardiomegaly without pulmonary edema.  No pleural
 effusions.}
         & {\small(b) \textit{Radiology findings after 13 hours}: There is mild bibasilar atelectasis.  There is no
 pneumothorax or large pleural effusion.}
         & {\small (c) CXR image generated by \ddlcxr given the initial CXR image shown in (a) and the EHR data within the 13 hours. }
    \end{tabular}
    }
     \caption{Case Study of Sample \#5}\label{fig:sample5}
\end{figure}

\begin{figure}[h!]\small
    \centering
    \scalebox{0.85}{
    \begin{tabular}{C{15em} C{15em} C{15em}} 
        \includegraphics[width=0.9\linewidth, valign=c]{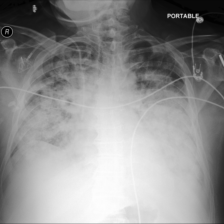}
        & \includegraphics[width=0.9\linewidth, valign=c]{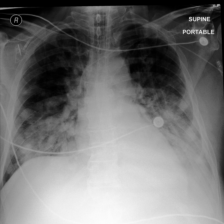}
        & \includegraphics[width=0.9\linewidth, valign=c]{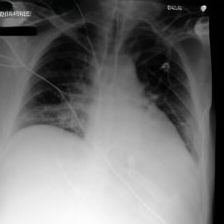}\\
    \end{tabular}
    }
    \scalebox{0.85}{
    \begin{tabular}{p{15em} p{15em} p{15em}}
        {\small(a) \textit{Initial radiology findings}: The lungs bilaterally demonstrate severe, extensive rounded nodular densities.
 }
         & {\small(b) \textit{Radiology findings after 31 hours}:  Bilateral nodular parenchymal opacities are
 unchanged in this patient with known lymphoma.  There are likely layering
 effusions, left greater than right.}
         & {\small (c) \small CXR image generated by \ddlcxr given the initial CXR image shown in (a) and the EHR data within the 31 hours.}
    \end{tabular}
    }
     \caption{Case Study of Sample \#6}\label{fig:sample6}
\end{figure}

\begin{figure}[h!]\small
    \centering
    \scalebox{0.85}{
    \begin{tabular}{C{15em} C{15em} C{15em}} 
        \includegraphics[width=0.9\linewidth, valign=c]{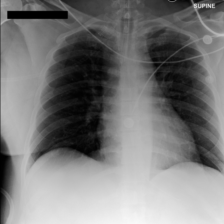}
        & \includegraphics[width=0.9\linewidth, valign=c]{visual_examples2/50__28-29/50__28-29_x1.png}
        & \includegraphics[width=0.9\linewidth, valign=c]{visual_examples2/50__28-29/50__28-29_gen.png}\\
    \end{tabular}
    }
    \scalebox{0.85}{
    \begin{tabular}{p{15em} p{15em} p{15em}}
        {\small(a)\textit{Initial radiology findings}:  Heart size is normal.  Mediastinal
 and hilar contours are within normal limits.  Pulmonary vasculature is normal.

 }
         & {\small(b) \textit{Radiology findings after 29 hours}: Small amount of right pleural effusion is present, more conspicuous than on
 the prior study.}
         & {\small (c) \small CXR image generated by \ddlcxr given the initial CXR image shown in (a) and the EHR data within the 29 hours.}
    \end{tabular}
    }
     \caption{Case Study of Sample \#7}\label{fig:sample7}
\end{figure}

\end{document}